\documentclass[preprint,12pt]{elsarticle}

\usepackage{amsthm, amsmath, amssymb, bm}
\usepackage{algorithm2e}
\usepackage{natbib}
\usepackage{graphicx}
\usepackage{xcolor}
\usepackage{tikz}
\usetikzlibrary{arrows.meta, calc, positioning}
\usepackage{booktabs}
\usepackage[colorlinks=true,
            linkcolor=blue,
            urlcolor=blue]{hyperref}
\journal{Journal of Computational Physics}

\theoremstyle{plain}

\newtheorem{Proposition}{Proposition}
\theoremstyle{remark}
\newtheorem{Remark}{Remark}

\newcommand{\cX}{\mathcal{X}}

\newcommand{\R}{\mathbb{R}}

\newcommand{\stepPOD}{\textcircled{\raisebox{-0.4pt}{\footnotesize 1}}}
\newcommand{\stepPERT}{\textcircled{\raisebox{-0.4pt}{\footnotesize 2}}}
\newcommand{\stepGP}{\textcircled{\raisebox{-0.4pt}{\footnotesize 3}}}
\newcommand{\stepTRANSP}{\textcircled{\raisebox{-0.4pt}{\footnotesize 4}}}
\newcommand{\stepVAR}{\textcircled{\raisebox{-0.4pt}{\footnotesize 5}}}
\newcommand{\stepCRC}{\textcircled{\raisebox{-0.4pt}{\footnotesize 6}}}

\definecolor{stepgray}{HTML}{F1EFE8}   \definecolor{stepgrayL}{HTML}{5F5E5A}
\definecolor{stepcoral}{HTML}{FAECE7}  \definecolor{stepcoralL}{HTML}{993C1D}
\definecolor{stepteal}{HTML}{E1F5EE}   \definecolor{steptealL}{HTML}{0F6E56}
\definecolor{steppurple}{HTML}{EEEDFE} \definecolor{steppurpleL}{HTML}{534AB7}
\definecolor{stepblue}{HTML}{E6F1FB}   \definecolor{stepblueL}{HTML}{185FA5}

\begin{document}

\begin{frontmatter}

\title{Conformal risk control for model-form uncertainty in parametric non-intrusive reduced-order models}

\author[ENS]{Edgar Jaber}
\author[ENS,Michelin]{Rémy Vallot}
\author[ENS,Michelin]{Thibault Dairay}
\author[ENS,ENSIIE]{Mathilde Mougeot}

\affiliation[ENS]{organization={Université Paris-Saclay, CNRS, ENS Paris-Saclay, Centre Borelli},
            city={Gif-sur-Yvette},
            postcode={91190},
            country={France}}
\affiliation[Michelin]{organization={Manufacture Française des Pneumatiques Michelin},
            city={Clermont-Ferrand},
            postcode={63000},
            country={France}}
\affiliation[ENSIIE]{
            organization={ENSIIE},
            city={Évry-Courcouronnes},
            postcode={91000},
            country={France}}

\begin{abstract}
Non-intrusive reduced-order models (NIROMs) have become a standard tool for approximating parametric partial differential equations from computer design of experiments while significantly reducing computational costs. However, assessing the reliability of their predictions remains a major challenge, particularly in extrapolation regimes or under limited training data. In this work, we introduce a framework for quantifying model-form uncertainty in NIROMs by combining a perturbative stochastic representation of reduced bases with distribution-free conformal-type methods. Starting from a deterministic reduced basis constructed from snapshot matrices, we model uncertainty through random perturbations defined on the Stiefel manifold, directed along the discarded modes, yielding stochastic reduced-order approximations whose induced variance reflects the basis-truncation error. A transport approximation gives a closed-form posterior variance that separates basis-induced from regression-induced uncertainty, without re-training the underlying Gaussian processes. We include this posterior variance within a conformal risk control calibration framework, that provides prediction sets with coordinate miscoverage guarantees. The calibration factor produced by this framework is itself an interpretable, scalar diagnostic of the quality of the uncertainty estimate. The methodology is evaluated on parametric PDE benchmarks and an industrial tire-manufacturing calendering process. Numerical experiments demonstrate reliable, locally informative uncertainty quantification that goes beyond the Gaussian predictive variance.
\end{abstract}

\begin{highlights}
\item A framework to quantify model-form uncertainty in non-intrusive reduced-order models with Gaussian-process modes, through random Stiefel-manifold perturbations of a reference POD basis directed along the discarded modes.
\item A first-order transport approximation separates basis-induced (structural) uncertainty from regression-induced (data) uncertainty and yields a closed-form local parametric variance combining both.
\item The model-form variance is calibrated within a conformal risk control framework that generalizes split conformal prediction to multidimensional functional outputs with distribution-free guarantees on the expected coordinate miscoverage.
\item The conformal calibration factor quantifies the quality of the uncertainty estimate: it is small for the perturbative variance and an order of magnitude larger for the Gaussian posterior variance, demonstrating that the perturbative variance matches the true pointwise error.
\end{highlights}


\end{frontmatter}

\section{Introduction}
\label{sec1}
\noindent Reduced order models (ROMs) \citep{Quarteroni2016, Hesthaven2016, Benner2020} have emerged over the past two decades as a principled framework for accelerating the simulation of complex, computationally intensive parametric partial differential equations (PDEs) typically discretized through finite element, finite volume, or finite difference schemes. These surrogate models are designed to approximate high-fidelity solutions defined on high-resolution meshes while significantly reducing computational complexity by projecting the dynamics onto low-dimensional subspaces capturing the dominant features of the solution manifold. Broadly speaking, two main classes of ROMs can be distinguished: \emph{intrusive} approaches, which rely on Galerkin-type projections of the governing equations onto reduced bases derived via the reduced basis method \citep{Quarteroni2016, Hesthaven2016}, and \emph{non-intrusive} approaches, which are primarily data-driven and constructed from proper orthogonal decomposition (POD) of snapshot matrices without requiring access to the source code of the high-fidelity solver \citep{Hesthaven2018, Peherstorfer2016, Kramer2024}. The latter has become especially attractive in industrial settings where commercial or legacy solvers are treated as black boxes \citep{Geelen2023, McQuarrie2023, Jaber2026-thesis}.

\noindent A wide range of non-intrusive strategies have been developed in recent years. Operator Inference \citep{Peherstorfer2016, Kramer2024, McQuarrie2023} learns reduced operators by exploiting the polynomial structure inherited from the governing equations. POD-based regression approaches couple a linear reduced basis with regression in the latent space, using Gaussian processes \citep{Guo2018, Jaber2025-2}, neural networks \citep{Hesthaven2018, Fresca2022}, or kernel methods \citep{Salvador2021}. To overcome the Kolmogorov barrier inherent to linear approximation spaces, deep-learning-based ROMs employ nonlinear encoders and decoders \citep{Fresca2022, Lee2020, Franco2023}, while graph- and manifold-based approaches \citep{Pichi2024, Geelen2023} extend these ideas to unstructured geometries. Applications span a broad spectrum, including turbulent and incompressible flows \citep{Stabile2018, Ahmed2021}, fluid–structure interactions \citep{Azzeddine2024}, aeroelasticity and combustion in propulsion systems \citep{Farcas2023, Zastrow2025}, structural and degradation analysis in nuclear reactors \citep{Jaber2025-3}, and digital twins in energy and infrastructure systems \citep{Hartmann2018, Niederer2021}.

\noindent For many downstream applications of surrogate modeling, including  rapid decision support \citep{DeRocquigny2008}, uncertainty quantification \citep{Sullivan2015, Smith2024}, and Bayesian inverse problems \citep{Stuart2010}, assessing the predictive reliability of the reduced model is of central importance. The discrepancy between a ROM and its high-fidelity counterpart is commonly referred to as \emph{model-form uncertainty} (MFU) \citep{Soize2017-1}. By model-form uncertainty we mean, precisely, the component of the prediction error that is attributable to the \emph{structure} of the surrogate rather than to the finiteness of the training sample: the choice of truncation rank, the choice of reduced basis, and the choice of regression model in the latent space. It is to be distinguished from the estimation (or regression-induced) uncertainty, which quantifies the error committed when learning the reduced coordinates from a finite design of experiments, and which does vanish as the design of experiments is refined. Model-form uncertainty, by contrast, cannot be reduced by collecting more training data alone, since it stems from the structural choices made when constructing the reduced model itself \citep{Drohmann2015, Manzoni2021}. Separating these two contributions, and quantifying the first of them, is the object of the present work.

\noindent In the context of intrusive parametric ROMs, a perturbative framework was introduced by \citep{Soize2017-1} to quantify such uncertainty through stochastic reduced-order bases. The Stiefel manifold is the natural geometric object here since a POD basis is by construction a matrix with orthonormal columns, and the set of such matrices is the Stiefel manifold. Any probabilistic perturbation of the basis that is to yield a legitimate reduced-order model must therefore keep the perturbed basis on that manifold, since an oblique or rank-deficient basis would destroy the orthogonal-projection interpretation of the reduced coordinates and make the resulting variance decomposition meaningless. Perturbing on the manifold, rather than in the ambient matrix space followed by re-orthonormalization, is what guarantees that every realization of the stochastic basis is itself an admissible POD-type basis. Restricting the perturbation to suitable subsets of the manifold is what encodes the modelling knowledge: the perturbation is not allowed to explore all admissible bases, but only those obtained by tilting the retained modes toward the subspace spanned by the discarded ones, which is precisely where the truncation error resides. This construction is made explicit in Section~\ref{sec32}. The framework was extended in \citep{SoizeFarhat2019} to a probabilistic learning formulation in which the perturbation hyperparameters are identified through a statistical inverse problem, and generalizes earlier developments from structural dynamics, where stochastic ROMs were derived using dedicated random matrix constructions, notably maximum entropy formulations \citep{Soize2017}. Related a posteriori error indicators and residual-based estimators have also been proposed, but typically require intrusive access to the discretized operators \citep{Manzoni2021, Hesthaven2016}.\\

\noindent The present work extends the perturbative framework of \citep{Soize2017-1} to the non-intrusive setting of POD-driven surrogate models by introducing probabilistic perturbations of a reference POD basis. Adopting a Bayesian formulation based on Gaussian processes \citep{Rasmussen2006} for learning the reduced modes, we combine the basis-induced contribution with the natural notion of prediction uncertainty in the parameter space. In contrast to the inverse-problem identification in \citep{SoizeFarhat2019}, we replace the statistical identification of the perturbation hyperparameters by the calibration of a single perturbation amplitude fixed a priori in order to propagate a first-order approximation onto the perturbed Gaussian surrogates, and the resulting posterior standard-deviation is used in a conformal risk-control layer in order to provide prediction ensembles with distribution-free risk guarantees.\\

\noindent The uncertainty calibration layer itself deserves to be introduced in three stages. First, conformal prediction \citep{Vovk2005, Angelopoulos2021} is a distribution-free methodology that converts any point predictor, together with a held-out calibration sample, into prediction sets enjoying finite-sample marginal coverage, under the sole assumption that calibration and test data are exchangeable. It requires no assumption on the correctness of the underlying model, which makes it particularly attractive for surrogates whose error structure is unknown. Second, conformal risk control (CRC) \citep{Angelopoulos2024-CRC} generalizes this idea by replacing the binary notion of coverage with a user-specified expected loss: instead of guaranteeing that the true output falls in the prediction set with prescribed probability, CRC guarantees that a monotone risk functional of the prediction set stays below a prescribed level. This added flexibility is what makes the framework applicable to the multidimensional, field-valued outputs considered here, for which coordinate-wise coverage is the natural quantity to control. Finally third, adaptive variants of CRC \citep{Blot2025} locally adjust the ensemble widths to different notions of miscoverage; they are particularly well suited to ROMs, where predictive accuracy is heterogeneous in the parameter space. Conformal prediction has recently attracted considerable attention in the scientific machine learning community, with applications to physics-informed neural networks \citep{Yu2026}, latent-space ROMs \citep{Katona2025}, neural operators, Gaussian process interpolation \citep{Jaber2025-2}, and engineering surrogates \citep{Jaber2026-thesis}.\\

\noindent We restrict attention in this work to linear POD-based ROMs which we also call NIROMs (Non-intrusive reduced-order models). Extensions to nonlinear manifold-based ROMs (such as autoencoder-based methods) require a different treatment of basis perturbations and are left for future investigation. However, some recent work in this direction can be found in \citep{Yong2025}.

\paragraph{Contributions}
\noindent Our contributions can be summarized as follows:
\begin{enumerate}
    \item We propose a perturbative model-form uncertainty framework for Gaussian process NIROMs, combining a Stiefel-manifold perturbation of the POD basis with the posterior variance of the reduced coordinates;
    \item We derive a closed-form posterior covariance that decomposes the predictive uncertainty into a basis-induced (model-form) component and a regression-induced (data-driven) component, using a first-order transport approximation that avoids retraining the Gaussian processes for each perturbation realization;
    \item We wrap the perturbative posterior in a conformal risk-control framework that provides distribution-free prediction ensembles, and we show that the resulting calibration factor is itself an interpretable diagnostic of the quality of the underlying variance estimate;
    \item We demonstrate the methodology on four numerical examples: a 2D parametric Poisson problem, a 1D parametric linear advection equation, a 1D viscous Burgers equation, and an industrial 2D calendering process arising in tire manufacturing.
\end{enumerate}
The methodology is fully reproducible and based on standard Python scientific machine learning libraries, with code available in the following \href{https://github.com/EdgarJaber/MFU-NIROMs.git}{\texttt{GitHub repository}}.

\section{Notations}
\label{sec2}
\noindent We begin by introducing the computational model, and only then the parameters it depends on and the design of experiments built from it.\\

\noindent \textbf{Computational model.} Let:
\begin{equation}
u : \cX\subset \R^{d} \longrightarrow \R^{N},
\end{equation}
denote a computational model, typically arising from the discretization of a parametric PDE on a fixed spatial or space-time mesh. In other words, it can be seen as a numerical solution operator such that for a PDE problem $E_{\bm{\mu}}$, $u(\bm{\mu}) = \texttt{solve}(E_{\bm{\mu}})$. The integer $N$ is the number of degrees of freedom of that mesh, so that $u(\bm{\mu})\in\R^{N}$ is the discretized solution field, one entry per mesh node. Throughout, $N$ is large, and the cost of a single evaluation of $u$ is the bottleneck the surrogate is meant to remove.\\

\noindent \textbf{Input parameters.} The set $\cX\subset\R^{d}$ is the admissible parameter domain and $\bm{\mu} = (\mu_1,\ldots,\mu_d)^\top \in \cX$ the vector of input parameters of $u$, assumed to follow a probability distribution $\bm{\mu} \sim \pi_{\bm{\mu}}\in \mathcal{P}(\cX)$, where $\mathcal{P}(\cX)$ denotes the set of probability measures on $\cX$.\\

\noindent \textbf{Design of experiments.} We call:
\begin{equation}
\mathrm{DoE}^{u}_{\pi_{\bm{\mu}}} = \{(\bm{\mu}^{(i)}, u(\bm{\mu}^{(i)}))\}_{i=1}^n,
\qquad \bm{\mu}^{(i)} \sim \pi_{\bm{\mu}},
\end{equation}
a design of experiments of size $n$, i.e.\ a finite collection of parameter values together with the corresponding high-fidelity solutions. The associated snapshot matrix is:
\begin{equation}
\bm{Y} = \left[u(\bm{\mu}^{(1)}), \ldots, u(\bm{\mu}^{(n)})\right] \in \R^{N \times n}.
\end{equation}

\noindent \textbf{Reduced basis.} A reduced basis of dimension $r \leq \min(N,n)$, where $r$ is the retained truncation rank, is obtained from the singular value decomposition $\bm{Y} = \bm{\Phi}_* \bm{D} \bm{W}^\top$, where $\bm{\Phi}_* = (\varphi_1,\ldots,\varphi_r) \in \R^{N \times r}$ contains orthonormal modes satisfying $\bm{\Phi}_*^\top \bm{\Phi}_* = I_r$. Such a matrix belongs to the Stiefel manifold
\begin{equation}
\mathrm{St}(N,r) = \{V\in\R^{N\times r} : V^{\top}V = I_{r}\}.
\label{eq:stiefel}
\end{equation}
The tangent space at a point $x$ of a manifold $M$ is denoted $T_{x}M$.\\


\section{Methodology}
\label{sec3}

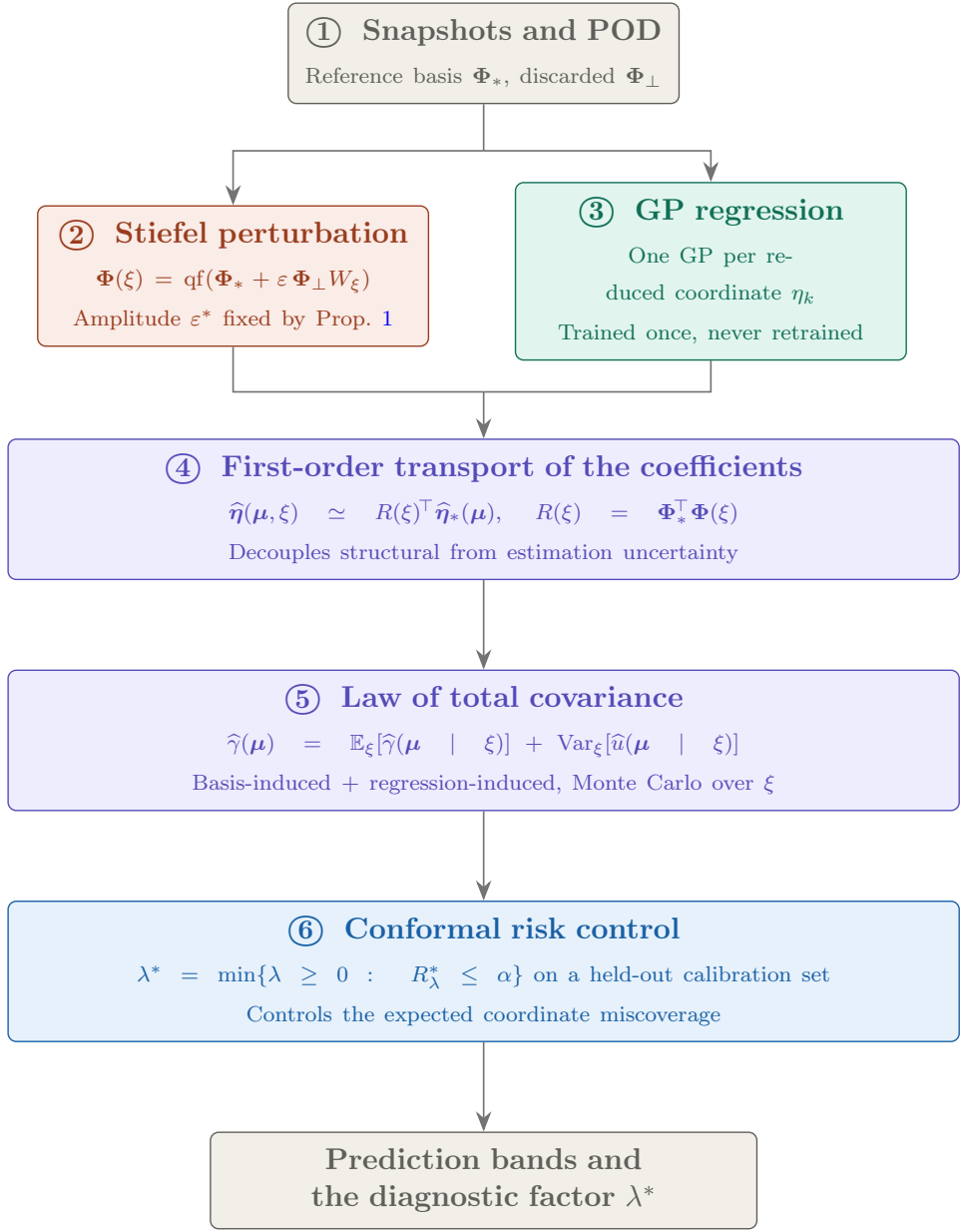
\begin{figure}[ht!]
\centering
\resizebox{0.92\textwidth}{!}{%
\begin{tikzpicture}[
  font=\small,
  sbox/.style={
    draw, rounded corners=3pt, line width=0.5pt,
    align=center, inner sep=6pt, text width=#1
  },
  snarrow/.style={sbox=4.6cm},
  swide/.style={sbox=11.8cm},
  sflow/.style={-{Stealth[length=2.6mm,width=2mm]}, line width=0.6pt, black!55},
  sline/.style={line width=0.6pt, black!55},
]

\node[snarrow, fill=stepgray, draw=stepgrayL] (pod) {
  {\color{stepgrayL}\small\bfseries \stepPOD\; Snapshots and POD}\\[2pt]
  {\color{stepgrayL}\scriptsize Reference basis $\bm{\Phi}_{*}$, discarded $\bm{\Phi}_{\perp}$}
};

\node[snarrow, fill=stepcoral, draw=stepcoralL,
      below left=1.3cm and 0.7cm of pod.south] (pert) {
  {\color{stepcoralL}\small\bfseries \stepPERT\; Stiefel perturbation}\\[2pt]
  {\color{stepcoralL}\scriptsize $\bm{\Phi}(\xi)=\mathrm{qf}(\bm{\Phi}_{*}+\varepsilon\,\bm{\Phi}_{\perp}W_{\xi})$}\\[1pt]
  {\color{stepcoralL}\scriptsize Amplitude $\varepsilon^{*}$ fixed by Prop.~\ref{prop:prop1}}
};

\node[snarrow, fill=stepteal, draw=steptealL,
      below right=1.0cm and 0.4cm of pod.south] (gp) {
  {\color{steptealL}\small\bfseries \stepGP\; GP regression}\\[2pt]
  {\color{steptealL}\scriptsize One GP per reduced coordinate $\eta_{k}$}\\[1pt]
  {\color{steptealL}\scriptsize Trained once, never retrained}
};

\node[swide, fill=steppurple, draw=steppurpleL,
      below=1.6cm of pod.south, anchor=north, yshift=-2.7cm] (transp) {
  {\color{steppurpleL}\small\bfseries \stepTRANSP\; First-order transport of the coefficients}\\[2pt]
  {\color{steppurpleL}\scriptsize $\widehat{\bm{\eta}}(\bm{\mu},\xi)\simeq R(\xi)^{\!\top}\widehat{\bm{\eta}}_{*}(\bm{\mu})$, \quad $R(\xi)=\bm{\Phi}_{*}^{\!\top}\bm{\Phi}(\xi)$}\\[1pt]
  {\color{steppurpleL}\scriptsize Decouples structural from estimation uncertainty}
};

\node[swide, fill=steppurple, draw=steppurpleL, below=1.15cm of transp] (cov) {
  {\color{steppurpleL}\small\bfseries \stepVAR\; Law of total covariance}\\[2pt]
  {\color{steppurpleL}\scriptsize $\widehat{\gamma}(\bm{\mu})=\mathbb{E}_{\xi}[\widehat{\gamma}(\bm{\mu}\mid\xi)]+\mathrm{Var}_{\xi}[\widehat{u}(\bm{\mu}\mid\xi)]$}\\[1pt]
  {\color{steppurpleL}\scriptsize Basis-induced $+$ regression-induced, Monte Carlo over $\xi$}
};

\node[swide, fill=stepblue, draw=stepblueL, below=1.15cm of cov] (crc) {
  {\color{stepblueL}\small\bfseries \stepCRC\; Conformal risk control}\\[2pt]
  {\color{stepblueL}\scriptsize $\lambda^{*}=\min\{\lambda\ge 0:\;R^{*}_{\lambda}\le\alpha\}$ on a held-out calibration set}\\[1pt]
  {\color{stepblueL}\scriptsize Controls the expected coordinate miscoverage}
};

\node[sbox=6.6cm, fill=stepgray, draw=stepgrayL, below=1.15cm of crc] (out) {
  {\color{stepgrayL}\small\bfseries Prediction bands and the diagnostic factor $\lambda^{*}$}
};

\coordinate (split) at ($(pod.south)+(0,-0.6)$);
\coordinate (merge) at ($(transp.north)+(0,0.6)$);

\draw[sline] (pod.south) -- (split);
\draw[sflow] (split) -| (pert.north);
\draw[sflow] (split) -| (gp.north);

\draw[sline] (pert.south) |- (merge);
\draw[sline] (gp.south)   |- (merge);
\draw[sflow] (merge) -- (transp.north);
\draw[sflow] (transp) -- (cov);
\draw[sflow] (cov) -- (crc);
\draw[sflow] (crc) -- (out);

\end{tikzpicture}%
}
\caption{%
    \textbf{Synoptic view of the proposed methodology.}
    Step \stepPOD{} (Section~\ref{sec31}) builds the reference POD basis $\bm{\Phi}_{*}$ and retains the discarded modes $\bm{\Phi}_{\perp}$. Steps \stepPERT{} (Section~\ref{sec32}) and \stepGP{} (Section~\ref{sec31}) are carried out independently: the basis is perturbed on the Stiefel manifold along the discarded modes, while the Gaussian processes are trained once and for all on the reference reduced coordinates. Step \stepTRANSP{} (Section~\ref{sec33}) is the key approximation that couples the two branches without retraining, and step \stepVAR{} (Section~\ref{sec33}) assembles the local variance by the law of total covariance. Step \stepCRC{} (Section~\ref{sec34}) calibrates a single scalar $\lambda^{*}$ against a held-out sample, yielding both the prediction band and a scalar diagnostic of variance quality.%
}
\label{fig:synoptic}
\end{figure}

\noindent The methodology proceeds in six successive steps, summarized in the synoptic diagram of Figure~\ref{fig:synoptic}. Each step involves a specific modeling choice, and the numbering introduced there is used consistently throughout the remainder of the paper. Steps \stepPOD-\stepGP{} set up the objects, steps \stepTRANSP-\stepVAR{} produce the local variance, and step \stepCRC{} turns that variance into a guaranteed prediction band.\\

\noindent For a costly parametric simulation code $u$, the main idea of this approach is to start from a parametric non-intrusive reduced-order model (NIROM) of the form:
\begin{equation}
    \widehat{u}(\bm{\mu}) = \bm{\Phi}_{*}\widehat{\bm{\eta}}_{*}(\bm{\mu}) = \sum_{k=1}^{r}\widehat{\eta}_{k}(\bm{\mu})\varphi_{k} \in \R^N,
\label{eq:nirom}
\end{equation}
and to build a notion of \emph{local variance} of the learning method based both on the structural uncertainty of the modes and on the regression error of the reduced coordinates. In order to quantify MFU for such a non-intrusive surrogate, we first address, in step \stepPOD, the construction of the basis $\bm{\Phi}_{*} = (\varphi_{1},\ldots,\varphi_{r})\in \R^{N\times r}$, which satisfies $\bm{\Phi}_{*}^{\top}\bm{\Phi}_{*}=I_{r}$. Such a matrix belongs to the Stiefel manifold $\mathrm{St}(N,r)$ defined in Eq.~\eqref{eq:stiefel} \citep{Absil2008}. The objective of step \stepPERT{} is to perturb the basis probabilistically around the reference modes $\bm{\Phi}_{*}$, while preserving the constraint that the perturbed basis remains on $\mathrm{St}(N,r)$. This perturbative approach aims at quantifying the non-intrusive model-form uncertainty stemming from the selection of relevant modes and the POD algorithm. In other words, we seek $\Delta_{\xi}$ such that $\bm{\Phi}(\xi) = \bm{\Phi}_{*} + \Delta_{\xi} \in \text{St}(N,r)$, where $\xi$ is a seed drawn from a probability space. From the perturbed basis, one obtains a new projection of the design of experiments and can re-learn the reduced coordinates, yielding a probabilistic NIROM:
 \begin{equation}
     \widehat{u}(\bm{\mu}, \xi) = \bm{\Phi}(\xi) \widehat{\bm{\eta}}(\bm{\mu},\xi)= \sum_{k=1}^{r}\widehat{\eta}_{k}(\bm{\mu},\xi)\varphi_{k}(\xi).
 \label{eq:naive_nirom}
 \end{equation}
 However, this naive approach is computationally costly: estimating statistics of the perturbations requires retraining the Gaussian processes for every realization of $\xi$, and the projected modes and perturbed basis are not probabilistically independent. We therefore seek a computationally tractable decoupling of these two sources of uncertainty. Denoting by $\omega$ an independent seed for the GP posterior, we consider a realization:
 \begin{equation}
     \widehat{u}(\bm{\mu}, \omega, \xi) = \bm{\Phi}(\xi) \widehat{\bm{\eta}}(\bm{\mu},\omega, \xi).
 \end{equation}
The simplification performed in step \stepTRANSP{} consists in approximating the GP prediction in the perturbed basis by a linear transport of the prediction in the reference basis, using a matrix $R(\xi)\in\R^{r\times r}$ such that $\widehat{\bm{\eta}}(\bm{\mu},\omega, \xi) \simeq R(\xi)^{\top}\widehat{\bm{\eta}}_{*}(\bm{\mu},\omega)$. With this transport, the prediction factors as:
\begin{equation}
    \widehat{u}(\bm{\mu}, \omega, \xi) \;\simeq\; \underbrace{\bm{\Phi}(\xi)R(\xi)^{\top}}_{\text{Structural uncertainty}} \underbrace{\widehat{\bm{\eta}}_{*}(\bm{\mu},\omega)}_{\text{Estimation uncertainty}},
\label{eq:factorization}
\end{equation}
which decouples the two uncertainty sources and avoids any GP retraining. This factorization is what makes the whole construction affordable, since it replaces GP trainings by matrix products. As we will see below, the methodology is dependent on a perturbation amplitude parameter $\varepsilon$ that controls the spread of the perturbation and whose calibration enables the preceding factorization. Once an estimate of the local variance is available from step \stepVAR, we use it in step \stepCRC{} inside a conformal risk-control procedure in order to obtain distribution-free prediction sets that quantify the model-form uncertainty.

\subsection{Steps $\textnormal{\stepPOD{}}$ and $\textnormal{\stepGP}$: non-intrusive reduced order models with Gaussian Processes}
\label{sec31}
\noindent Let $(u:\bm{\mu}\mapsto u(\bm{\mu}))$ be a computational PDE solver for a given parametric PDE problem. The solver is treated as a black box, in the sense that the numerical solver cannot be modified. We assume that the inputs have been modeled probabilistically with $\bm{\mu} \sim \pi_{\bm{\mu}}\in\mathcal{P}(\cX)$ and that we have access to a design of computer experiments $\text{DoE}_{\pi_{\bm{\mu}}}^{u} = \{(\bm{\mu}^{(i)},u(\bm{\mu}^{(i)}))\}_{i=1}^{n}$ following a Monte-Carlo sampling in $\pi_{\bm{\mu}}$. To simplify the presentation, the mean of $u$ over the realizations of $\bm{\mu}$ is assumed to be zero. We collect the snapshots in the matrix:
\begin{equation}
\bm{Y} = \left[ u(\bm{\mu}^{(1)}), \ldots, u(\bm{\mu}^{(n)}) \right] \in \mathbb{R}^{N \times n}.
\end{equation}
The idea of reduced-order modeling is to reduce the dimensionality of the output space by applying a Karhunen--Loève expansion (KLE) in this probabilistic setting \citep{LeMaitre2010, Sullivan2015}, which is computed in the discrete case like a proper orthogonal decomposition (POD) using the empirical covariance matrix:
\begin{equation}
\widehat{C} = \frac{1}{n} \bm{Y} \bm{Y}^\top.
\end{equation}
We then perform a singular value decomposition of $\bm{Y}$,
\begin{equation}
\bm{Y} = \bm{\Phi}_{*} \bm{D} \bm{W}^\top,
\end{equation}
where $\bm{\Phi}_{*} = (\varphi_1,\ldots,\varphi_r) \in \mathbb{R}^{N \times r}$ contains the orthonormal POD modes, $\bm{D} = \mathrm{diag}(\sigma_1,\ldots,\sigma_r)$ with $\sigma_1 \geq \cdots \geq \sigma_r \geq 0$, and $r \leq \min(N,n)$ is the number of retained modes. The truncation level $r$ is chosen so as to capture a prescribed proportion of the total variance (usually $99.9\%$). We denote by
\begin{equation}
    \bm{\Phi}_{\perp} =(\varphi_{r+1},\ldots,\varphi_{r+r_{\perp}})
\in \mathbb{R}^{N \times r_{\perp}}
\end{equation}
the matrix collecting additional $r_{\perp}\leq N-r$ discarded POD modes that will be used in the perturbation construction of Section~\ref{sec32}. By construction, $\bm{\Phi}_{*}^{\top}\bm{\Phi}_{\perp} = 0$ and $\bm{\Phi}_{\perp}^{\top}\bm{\Phi}_{\perp} = I_{r_{\perp}}$. Each snapshot $u(\bm{\mu}^{(i)})$ is then projected onto the retained POD modes:
\begin{equation}
\eta_k(\bm{\mu}^{(i)}) = \langle u(\bm{\mu}^{(i)}), \varphi_k \rangle = u(\bm{\mu}^{(i)})^\top \varphi_k,
\label{eq:kl_modes}
\end{equation}
yielding $r$ scalar-valued independent datasets:
\begin{equation}
\text{DoE}_k = \left\{ \left( \bm{\mu}^{(i)}, \eta_k(\bm{\mu}^{(i)}) \right) \right\}_{i=1}^n, \quad k = 1,\ldots,r.
\end{equation}
In step \stepGP, for each mode $k=1,\ldots,r$, we learn the mapping $(\bm{\mu} \mapsto \eta_k(\bm{\mu}))$ using a Gaussian process model \citep{Rasmussen2006}, yielding a posterior such that $\bm{\eta}_{*}(\bm{\mu})\sim \mathcal{N}(\widehat{\bm{\eta}}_{*}(\bm{\mu}),\widehat{\bm{\Sigma}}_{*}(\bm{\mu}))$, with $\widehat{\bm{\Sigma}}_*(\bm\mu)$ diagonal by independence of the $r$ scalar GPs. Details on Gaussian process regression are recalled in~\ref{app:gp}. This step is performed once and never repeated. The reduced-order mean predictor of the solution is then Eq.~\eqref{eq:nirom}.

\subsection{Step $\textnormal{\stepPERT}$: random perturbation of the modes}
\label{sec32}
\noindent Let $\bm{\Phi}_{*} \in \mathbb{R}^{N \times r}$ be the matrix of the reference POD basis, so that $\bm{\Phi}_{*}^\top \bm{\Phi}_{*} = I_r$ and $\bm{\Phi}_{*}\in\mathrm{St}(N,r)$. In order to introduce a small perturbation that preserves orthonormality, one constructs a perturbation in the tangent space of the Stiefel manifold at $\bm{\Phi}_{*}$, as proposed in \citep{Soize2017-1, SoizeFarhat2019}. The tangent space is given by \citep{Absil2008, Chikuse2003}:
\begin{equation}
T_{\bm{\Phi}_{*}} \mathrm{St}(N,r) = \left\{ \Delta \in \mathbb{R}^{N \times r} \;\middle|\; \bm{\Phi}_{*}^\top \Delta + \Delta^\top \bm{\Phi}_{*} = 0 \right\}.
\end{equation}
Given a random matrix $Z_{\xi} \in \mathbb{R}^{N \times r}$, a tangent perturbation is obtained by projecting $Z_{\xi}$ onto $T_{\bm{\Phi}_{*}} \mathrm{St}(N,r)$:
\begin{equation}
\Delta_{\xi} = Z_{\xi} - \bm{\Phi}_{*} \,\mathrm{sym}(\bm{\Phi}_{*}^\top Z_{\xi}),
\qquad \text{with} \quad \mathrm{sym}(A) = \tfrac{1}{2}(A + A^\top).
\end{equation}
This construction ensures $\Delta_{\xi} \in T_{\bm{\Phi}_{*}} \mathrm{St}(N,r)$. A perturbed basis is then defined by the following retraction:
\begin{equation}
\bm{\Phi}(\xi) = \mathrm{qf}(\bm{\Phi}_{*} + \varepsilon \Delta_{\xi}),
\label{eq:retraction}
\end{equation}
where $\varepsilon > 0$ controls the perturbation amplitude and $\mathrm{qf}(\cdot)$ denotes the orthonormal factor of a $QR$ decomposition \citep{Golub2013}. Specifically, if $\bm{\Phi}_{*} + \varepsilon \Delta_{\xi} = Q R$ is a $QR$ factorization, then:
\begin{equation}
\bm{\Phi}(\xi) = Q D,
\label{eq:perturbation_basis}
\end{equation}
where $D = \mathrm{diag}(\mathrm{sign}(\mathrm{diag}(R)))$ is selected so that perturbations vary smoothly and are not affected by sign discontinuities. By construction, $\bm{\Phi}(\xi) \in \mathrm{St}(N,r)$ and represents a small, random perturbation of $\bm{\Phi}_{*}$ that remains on the Stiefel manifold \citep{Absil2008}.

\subsection*{Selection of the random matrix}
\noindent A natural first option is to take the entries of $Z_{\xi}\in\mathbb{R}^{N\times r}$ as independent standard normal random variables, $Z_\xi \sim \mathcal{N}(0,1)^{\otimes N\times r}$. The projected perturbation:
\begin{equation}
\Delta_{\xi} = Z_{\xi} - \bm{\Phi}_{*} \,\mathrm{sym}(\bm{\Phi}_{*}^\top Z_{\xi})
\end{equation}
is then a centered Gaussian random element in $T_{\bm{\Phi}_{*}} \mathrm{St}(N,r)$, isotropic with respect to the Euclidean inner product on that space: for any orthogonal transformation $Q$ preserving the tangent space, the distributions of $\Delta_\xi$ and $Q\Delta_\xi$ coincide. This ambient-isotropic perturbation is directionally unbiased within $T_{\bm{\Phi}_{*}}\mathrm{St}(N,r)$, but for a truncated-basis ROM it carries no structural information: it explores all tangent directions equally, including those orthogonal to the truncation-error subspace, and the resulting induced field variance is spatially structureless. We therefore direct the perturbation along the discarded modes. Taking $W_{\xi} \sim \mathcal{N}(0,1)^{\otimes r_{\perp} \times r}$, we set:
\begin{equation}
Z_{\xi} = \bm{\Phi}_{\perp}\,W_{\xi} \in \mathbb{R}^{N \times r},
\end{equation}
which, by orthogonality $\bm{\Phi}_{*}^{\top}\bm{\Phi}_{\perp} = 0$, gives:
\begin{equation}
\Delta_{\xi} = \bm{\Phi}_{\perp}\,W_{\xi}  - \bm{\Phi}_{*}\,\mathrm{sym}(\bm{\Phi}_{*}^{\top}\bm{\Phi}_{\perp}W_{\xi}) = \bm{\Phi}_{\perp}\,W_{\xi},
\end{equation}
so that $\Delta_\xi$ already lies in $T_{\bm{\Phi}_{*}}\mathrm{St}(N,r)$ and requires no further projection. The perturbation thus rotates the retained basis toward the discarded subspace $\mathrm{span}(\bm{\Phi}_{\perp})$, which by construction contains the dominant part of the basis-truncation error. The induced field variance consequently concentrates where the discarded modes carry energy, giving the local variance a physically meaningful spatial structure aligned with the model-form error.\\

\noindent The transport matrix between the perturbed and reference bases is defined by:
\begin{equation}
R_{\varepsilon}(\xi) = \bm{\Phi}_{*}^{\top}\,\bm{\Phi}(\xi) \in \mathbb{R}^{r\times r}.
\label{eq:transport_matrix}
\end{equation}
It quantifies how far the perturbed basis is rotated away from the reference one. The following proposition shows that the expected deviation of $R_{\varepsilon}(\xi)$ from the identity is controlled at fourth order in $\varepsilon$, and can therefore be driven below any prescribed tolerance. This is what allows the amplitude $\varepsilon$ to be fixed a priori by the user, from an interpretable tolerance, rather than identified through an inverse problem as in \citep{SoizeFarhat2019}.

\begin{Proposition}
\label{prop:prop1}
Let $\bm{\Phi}(\xi) = \mathrm{qf}\!\left(\bm{\Phi}_{*} + \varepsilon\, \bm{\Phi}_{\perp} W_{\xi}\right)$ be the discarded-mode Stiefel perturbation, with $\bm{\Phi}_{*}^{\top}\bm{\Phi}_{\perp} = 0$ and $W_{\xi}\in\mathbb{R}^{r_{\perp}\times r}$. Then the overlap matrix admits the expansion:
\begin{equation}
R_{\varepsilon}(\xi)
= I_r - \tfrac{1}{2}\,\varepsilon^{2}\, W_{\xi}^{\top}W_{\xi}
  + O(\varepsilon^{4}),
\label{eq:overlap_expansion}
\end{equation}
and consequently:
\begin{equation}
\mathbb{E}_{\xi}\!\left\lVert R_{\varepsilon}(\xi) - I_r \right\rVert_{F}
= \tfrac{1}{2}\,\varepsilon^{2}\,
  \mathbb{E}_{\xi}\!\left\lVert W_{\xi}^{\top}W_{\xi} \right\rVert_{F}
  + O(\varepsilon^{4}).
\label{eq:overlap_expectation}
\end{equation}
In particular, for any tolerance $\tau > 0$, setting:
\begin{equation}
\varepsilon^{*}
= \left(
    \frac{2\tau}
         {\mathbb{E}_{\xi}\!\left\lVert W_{\xi}^{\top}W_{\xi}\right\rVert_{F}}
  \right)^{1/2},
\label{eq:eps_star}
\end{equation}
yields $\mathbb{E}_{\xi}\lVert R_{\varepsilon^{*}}(\xi) - I_r \rVert_{F} = \tau$
up to a term of order $O(\varepsilon^{4})$.
\end{Proposition}

\noindent The proof is provided in~\ref{app:proof1}. In the rest of the paper we will simply drop the subscript on the overlap matrix, writing $R(\xi) = R_{\varepsilon^{*}}(\xi)$.

\subsection{Steps $\textnormal{\stepTRANSP}$ and $\textnormal{\stepVAR}$: posterior MFU local variance}
\label{sec33}
\noindent Based on the previous construction, we have that $\bm{\Phi}(\xi)^{\top}\bm{\Phi}_{*}\simeq I_r$ in $\R^{r\times r}$ up to a term in $\varepsilon$ of order $4$. As mentioned in Section~\ref{sec3}, re-optimizing the GPs at each perturbation realization is prohibitively costly given that we want to build Monte Carlo estimations of the posterior local covariance. By using the calibrated $\varepsilon^{*}$ both bases span nearly the same subspace, and the reduced coordinates satisfy the transport relation of step \stepTRANSP:
\begin{equation}
\bm{\widehat{\eta}}(\bm{\mu},\xi) \simeq R(\xi)^\top \bm{\widehat{\eta}}_{*}(\bm{\mu}).
\label{eq:transport}
\end{equation}
This relation can be interpreted as a first-order transport of the reduced coordinates under a perturbation of the basis, and avoids retraining the GPs while preserving consistency with the underlying reduced space. By Proposition~\ref{prop:prop1}, the relation~\eqref{eq:transport} holds up to a residual of order $O(\varepsilon^{4})$. Therefore, all expressions derived from it in the remainder of this section inherit the same order of approximation, which we do not repeat at each step.\\

\noindent Let $\bm{\eta}_{*}(\bm{\mu})\sim \mathcal{N}(\widehat{\bm{\eta}}_{*}(\bm{\mu}),\widehat{\bm{\Sigma}}_{*}(\bm{\mu}))$ denote the GP posterior for the coefficients in the reference POD basis. Under the transport approximation, the dependence on $\xi$ can be viewed as a conditional dependence, giving the conditional GP posterior in the perturbed basis:
\begin{equation}
\widehat{\bm{\eta}}(\bm{\mu} \mid \xi) = R(\xi)^\top \widehat{\bm{\eta}}_{*}(\bm{\mu}),
\qquad
\widehat{\bm{\Sigma}}(\bm{\mu} \mid \xi) = R(\xi)^\top \widehat{\bm{\Sigma}}_{*}(\bm{\mu})\, R(\xi).
\end{equation}
The corresponding conditional prediction in the output ROM is:
\begin{equation}
\widehat{u}(\bm{\mu} \mid \xi)  = \bm{\Phi}(\xi)\, \widehat{\bm{\eta}}(\bm{\mu} \mid \xi) = \bm{\Phi}(\xi)R(\xi)^\top \widehat{\bm{\eta}}_{*}(\bm{\mu}),
\end{equation}
with conditional covariance:
\begin{equation}
\widehat{\gamma}(\bm{\mu} \mid \xi)
= \bm{\Phi}(\xi)\widehat{\bm{\Sigma}}(\bm{\mu} \mid \xi )\bm{\Phi}(\xi)^\top.
\end{equation}
In step \stepVAR, the total predictive uncertainty is obtained by marginalizing over the basis perturbations $\xi$. Using the law of total covariance:
\begin{equation}
\widehat{\gamma}(\bm{\mu}) = \mathbb{E}_{\xi}\!\left[ \widehat{\gamma}(\bm{\mu} \mid \xi) \right] + \mathrm{Var}_\xi\!\left[ \widehat{u}(\bm{\mu} \mid \xi) \right] \in \R^{N\times N},
\label{eq:total_variance}
\end{equation}
which expands to:
\begin{equation}
    \widehat{\gamma}(\bm{\mu}) = \mathbb{E}_{\xi}\!\left[\bm{\Phi}(\xi)R(\xi)^\top \widehat{\bm{\Sigma}}_{*}(\bm{\mu})\, R(\xi)\bm{\Phi}(\xi)^\top \right] + \mathrm{Var}_{\xi}\!\left[\bm{\Phi}(\xi)R(\xi)^\top \widehat{\bm{\eta}}_{*}(\bm{\mu})\right].
\label{eq:total_variance_expanded}
\end{equation}
The first term represents the average GP uncertainty propagated through the perturbed bases, while the second term captures the variability of the conditional mean induced by structural perturbations. Therefore the resulting averaged surrogate model is not a GP anymore. This decomposition naturally accounts for both statistical learning error and model-form uncertainty, without assuming independence between the two, and remains fully non-intrusive and computationally tractable through the transport approximation. In the rest of this section we will denote the standard-deviation as:
\begin{equation}
\widehat{\sigma}(\bm{\mu}):= (\widehat{\gamma}^{1/2}_{11}(\bm{\mu}), \ldots,\widehat{\gamma}^{1/2}_{NN}(\bm{\mu}))^{\top}\in\R^N.
\label{eq:standard_dev}
\end{equation}

\subsection*{Monte Carlo estimation}
\noindent In practice, the expectations with respect to $\xi$ in~\eqref{eq:total_variance}--\eqref{eq:total_variance_expanded} are approximated by Monte Carlo sampling. Given $M$ independent perturbations $\{\bm{\Phi}(\xi^{(m)})\}_{m=1}^M$, one computes
\begin{align}
\widehat{u}(\bm{\mu} \mid \xi^{(m)}) &= \bm{\Phi}(\xi^{(m)})R(\xi^{(m)})^{\top} \widehat{\bm{\eta}}_{*}(\bm{\mu}), \\
\widehat{\gamma}(\bm{\mu}\mid\xi^{(m)}) &= \bm{\Phi}(\xi^{(m)})R(\xi^{(m)})^{\top} \widehat{\bm{\Sigma}}_{*}(\bm{\mu})\, R(\xi^{(m)}) \bm{\Phi}(\xi^{(m)})^{\top}.
\end{align}
The total variance is then estimated as
\begin{equation}
\widehat{\gamma}(\bm{\mu})
\approx
\frac{1}{M} \sum_{m=1}^M \widehat{\gamma}(\bm{\mu}\mid\xi^{(m)})
+
\frac{1}{M} \sum_{m=1}^M
\left( \widehat{u}(\bm{\mu} \mid \xi^{(m)}) - \overline{u}(\bm{\mu}) \right)
\left(\widehat{u}(\bm{\mu} \mid \xi^{(m)}) - \overline{u}(\bm{\mu}) \right)^\top,
\label{eq:mfu_variance_est}
\end{equation}
where $\overline{u}(\bm{\mu}) = \frac{1}{M}\sum_{m=1}^{M}\widehat{u}(\bm{\mu} \mid \xi^{(m)})$ denotes the empirical mean over perturbations.

\subsection{Step $\textnormal{\stepCRC}$: conformal risk control prediction sets for multidimensional outputs}
\label{sec34}

\noindent Since the output of the simulation code is a discretized field,
\begin{equation}
    u(\bm{\mu}) = (u_{1}(\bm{\mu}),\ldots,u_{N}(\bm{\mu}))^{\top}\in\mathbb{R}^{N},
\end{equation}
constructing prediction sets requires handling the multidimensional nature of the output jointly. We therefore introduce in step \stepCRC{} a conformal risk control framework, which we adopt as our primary method for obtaining prediction bands guided by the MFU-informed local variance in Eq.~\eqref{eq:total_variance_expanded}.\\

\noindent Before describing the construction, we make explicit the property of split-conformal prediction that motivates the choice of CRC.

\begin{Remark}[Normalizer invariance of split conformal prediction]
\label{rem:invariance}
Consider mesh-wise split-conformal intervals \citep{Vovk2005} built from normalized scores $s^{(j)}_i = |u_i(\bm{\mu}^{(j)}) - \widehat{u}_i(\bm{\mu}^{(j)})| / \widehat{\sigma}_i(\bm{\mu}^{(j)})$, and let $q_{1-\alpha}$ denote their empirical $(1-\alpha)$-quantile over the calibration set. Replacing the normalizer $\widehat{\sigma}$ by $c\,\widehat{\sigma}$ for any constant $c>0$ rescales every score by $1/c$, hence rescales $q_{1-\alpha}$ by $1/c$, and leaves the resulting intervals $[\widehat{u}_i \pm q_{1-\alpha}\widehat{\sigma}_i]$ unchanged. The calibrated bands are therefore invariant under any global rescaling of the local variance, so the \emph{scale} of $\widehat{\sigma}$ cannot be assessed from them. Mahalanobis-score adaptive sets are too conservative and suffer from the same caveat.
\end{Remark}

\noindent Remark~\ref{rem:invariance} is the reason a different calibration layer is needed: we want a procedure in which the quality of the standard-deviation is directly observable. For this, we adopt \emph{conformal risk control} (CRC) \citep{Angelopoulos2024-CRC, Blot2025}, which calibrates a set-valued predictor to control a user-specified expected loss rather than a coverage probability.\\

\noindent For a loss functional
$\mathcal{L}:\mathbb{R}^{N}\times 2^{\mathbb{R}^{N}}\to[0,1]$ and a parametric family of sets $\{\Gamma_\lambda\}_{\lambda\ge 0}$ indexed by a scale parameter $\lambda\ge 0$, CRC selects a scale $\lambda^*$ enforcing:
\begin{equation}
\mathbb{E}\!\left[ \mathcal{L}\!\left(u(\bm{\mu}),\,\Gamma_{\lambda^*}(\bm{\mu})\right)
\right] \le \alpha.
\end{equation}
For the discretized-field outputs considered here we choose the average coordinate miscoverage loss,
\begin{equation}
\mathcal{L}(y,\Gamma)
= \frac{1}{N}\sum_{i=1}^{N}
  \bm{1}\!\left\{y_i \notin \mathrm{pr}_i(\Gamma)\right\},
\label{eq:crc_loss}
\end{equation}
where $y=(y_1,\ldots,y_N)^{\top}$ is an $N$-dimensional vector and $\mathrm{pr}_i(\Gamma) = \{y_i : y\in\Gamma\}\subset\R$ denotes the projection of the set $\Gamma\subset\R^{N}$ onto its $i$-th coordinate. This loss equals the fraction of mesh points not covered by $\Gamma$, so controlling it at level $\alpha$ guarantees that, on average, at most a fraction $\alpha$ of the field coordinates lie outside the band.\\

\noindent We consider the single-parameter family $\{\Gamma_{\lambda}\}_{\lambda\ge0}$ obtained by scaling the local MFU standard deviation defined in Eq.~\eqref{eq:standard_dev}:
\begin{equation}
\Gamma_{\lambda}(\bm{\mu})
= \prod_{i=1}^{N}
  \left[
    \widehat{u}_i(\bm{\mu})
    \pm
    \lambda\,\widehat{\sigma}_i(\bm{\mu})
  \right], \qquad \lambda \ge 0,
\label{eq:crc_band}
\end{equation}
the product denoting the Cartesian product of the $N$ coordinate-wise intervals. The parameter $\lambda$ is calibrated with the help of an independent \emph{calibration} dataset $\mathcal{D}_{\mathrm{cal}} = \{(\bm{\mu}^{(i)},u(\bm{\mu}^{(i)}))\}_{i=1}^{\ell}$ of size $\ell$. This dataset is independent from the training set used for training the initial surrogate. The empirical risk at scale $\lambda$ is defined by:
\begin{equation}
R_\lambda = \frac{1}{\ell}
  \sum_{j=1}^{\ell} \mathcal{L}\!\left(u(\bm{\mu}^{(j)}),\,\Gamma_\lambda(\bm{\mu}^{(j)})\right),
\end{equation}
and the conformal correction of \citep{Angelopoulos2024-CRC} is applied:
\begin{equation}
R^{*}_\lambda
= \frac{\ell}{\ell+1}\,R_\lambda + \frac{1}{\ell+1}.
\label{eq:crc_corrected}
\end{equation}
Since $R_\lambda$ is non-increasing in $\lambda$, the calibrated factor:
\begin{equation}
\lambda^{*} = \min\!\left\{\lambda \ge 0 : R^{*}_\lambda \le \alpha\right\},
\label{eq:lambda_star}
\end{equation}
is found by a monotone line search over a finite $\lambda$-grid. Under exchangeability of the calibration and test samples, the resulting band $\Gamma_{\lambda^*}$ satisfies the finite-sample
guarantee:
\begin{equation}
\mathbb{E}_{\bm{\mu}\sim\pi_{\bm{\mu}}}\!\left[\mathcal{L}\!\left(u(\bm{\mu}),\,\Gamma_{\lambda^*}(\bm{\mu})\right)
\right] \le \alpha,
\end{equation}
which translates to a distribution-free control of the expected fraction of out-of-band coordinates at level $\alpha$.\\

\noindent Because $\lambda^*$ is a single scalar shared across all mesh points and all parameter values, the relative shape of the band is fixed entirely by $\widehat{\sigma}_i(\bm{\mu})$ thus calibration sets only its overall scale. Two consequences follow. First, the normalizer invariance of Remark~\ref{rem:invariance} no longer holds: a local variance whose spatial profile matches the true error requires a smaller $\lambda^*$ and produces tighter bands at the same risk level, so the choice of $\widehat{\sigma}$ becomes both consequential and measurable. Second, $\lambda^*$ itself serves as a scalar diagnostic of uncertainty quality: a value of order unity indicates that the unscaled local variance already matches the true pointwise error magnitude, whereas a large value indicates systematic underestimation requiring strong corrective inflation. We report $\lambda^*$ for each benchmark and each choice of normalizer in Section~\ref{sec4}, where the perturbative MFU variance is shown to yield calibration factors within an order of magnitude of unity, while the Gaussian posterior variance yields $\lambda^* \gg 1$.\\

\noindent Finally, it is worth noting that although $\lambda^*$ is fixed, the band $\Gamma_{\lambda^*}$ remains adaptive across the parameter space through the spatial and parametric variation of $\widehat{\sigma}_i(\bm{\mu})$, realized band widths vary with $\bm{\mu}$. The construction is not adaptive in the stronger conditional sense, since $\lambda^*$ does not depend on the test parameter. Extensions that calibrate a parameter-dependent factor $\lambda(\bm{\mu},\bm{X})$ \citep{Blot2025}, or separate factors for the lower and upper bounds, would relax this restriction and are left for future work.

\section{Numerical results}
\label{sec4}

\noindent This section is organized as follows. In Section~\ref{sec40}, we first describe the models under study and the performance indicators used to compare them. We then report results on three academic benchmarks of increasing difficulty in Section~\ref{sec4acad}, for which a closed-form or highly resolved reference solution is available and the error budget can be attributed unambiguously. Finally, Section~\ref{sec44} is devoted to the industrial use case, where the method is deployed on a real tire-manufacturing process.

\subsection{Models and performance indicators}
\label{sec40}

\noindent \textbf{Models.} The three academic benchmarks are the following. The first is a 2D parametric Poisson problem with a moving source, a smooth benchmark whose solution manifold is highly compressible. The second is a 1D parametric linear advection equation on a periodic domain, where the solution manifold has poor linear-subspace approximability (slow Kolmogorov $r$-width decay), providing a challenging setting for a POD-based NIROM. The third is a 1D viscous Burgers equation, which introduces nonlinearity through the convective term and tests the methodology in a regime where the linear-POD assumption is stressed by the formation of thin fronts. Together they span the two extreme regimes of the error budget: regression-dominated (Poisson) and truncation-dominated (advection), with Burgers interpolating between them. \\
The industrial use case is a 2D rubber calendering process, illustrating the deployability of the method in a real engineering setting.\\

\noindent \textbf{Performance indicators.} For each model we report the following indicators:
\begin{enumerate}
    \item cross-sections of the predicted and reference fields together with their pointwise absolute error, assessing the quality of the underlying NIROM produced by steps \stepPOD-\stepGP;
    \item the CRC empirical risk as a function of the nominal level $1-\alpha$, which verifies that the guarantee of step \stepCRC{} is attained, together with the corresponding mean band width, which measures its efficiency;
    \item the calibration factor $\lambda^*$ of Eq.~\eqref{eq:lambda_star}, which measures how far the uncalibrated variance of step \stepVAR{} is from the true error scale, a value of order unity being the target;
    \item the Spearman correlation between the local band width and the absolute prediction error, which measures whether the variance is spatially informative, i.e. whether it is large where the error is large;
\end{enumerate}

\noindent As a baseline we compare with the pure-GP credibility interval \citep{Rasmussen2006}, i.e.\ the normalizer obtained by skipping steps \stepPERT{} and \stepTRANSP-\stepVAR{} and retaining only the Gaussian process posterior variance. Given the independence of the GP-learned modes, the posterior field variance is:
\begin{equation}
\widehat\sigma^{\mathrm{GP}}(\bm\mu)^2 = \sum_{k=1}^{r} \widehat\sigma_k(\bm\mu)^2\,\varphi_{*,k}^2 .
\label{eq:gp_normalizer}
\end{equation}
This comparison isolates the contribution of the basis perturbation: both normalizers are wrapped in the same CRC layer, so any difference in the calibrated bands is attributable to the variance model alone.\\

\noindent Table~\ref{tab:summary} collects, for each benchmark and each normalizer, the mean predictivity coefficient $\overline{R^2}$ over the test set, the mean absolute pointwise error $\mathrm{MAE}$, and the CRC calibration factor $\lambda^*$ at nominal level $1-\alpha = 0.90$. The ratio $\lambda^*_{\mathrm{GP}}/\lambda^*_{\mathrm{MFU}}$ is the central diagnostic of the paper: a value of order one for the MFU normalizer indicates that the perturbative variance already matches the true error scale before any calibration, whereas the large GP ratio confirms systematic underestimation of the model-form error.

\begin{table}[ht!]
\centering
\footnotesize
\caption{%
    \textbf{Summary of surrogate quality and CRC calibration factors.} Mean $R^2$ and mean absolute pointwise error $\mathrm{MAE}$ are computed on the test set $\mathcal{D}_{\mathrm{test}}$. The CRC calibration factor $\lambda^*$ is reported for both the perturbative MFU normalizer and the pure-GP normalizer at nominal level $1-\alpha = 0.90$, $\lambda^*_{\mathrm{GP}}/\lambda^*_{\mathrm{MFU}}$ quantifies the scale mismatch of the GP posterior variance relative to the true prediction error.%
}
\label{tab:summary}
\setlength{\tabcolsep}{8pt}
\renewcommand{\arraystretch}{1.25}
\begin{tabular}{lcccccc}
\toprule
\textbf{Benchmark}
  & $r$
  & $\overline{R^2}$
  & $\mathrm{MAE}$
  & $\lambda^*_{\mathrm{MFU}}$
  & $\lambda^*_{\mathrm{GP}}$
  & $\lambda^*_{\mathrm{GP}}/\lambda^*_{\mathrm{MFU}}$ \\
\midrule
Poisson (2D) & 13 & 0.9932 & $1.08\times 10^{-4}$ & 2.060 & 8.546 & $4.1\times$ \\
Advection (1D) & 3 & 0.3531 & $6.23\times 10^{-2}$ & 38.765 & 155.066 & $4.0\times$ \\
Burgers (1D) & 3 & 0.8491 & $1.76\times 10^{-2}$ & 15.317 & 61.911 & $4.0\times$ \\
Calendering (2D)   &  2 & 0.9993 & $1.06\times10^{4}$ & 1.141 & 3.128 & $2.7\times$ \\
\bottomrule
\end{tabular}
\end{table}

\noindent Throughout this section, the GPs use a constant mean and an RBF kernel, with hyperparameters estimated by maximum likelihood in GPyTorch; the corresponding code is available in the \href{https://github.com/EdgarJaber/MFU-NIROMs.git}{\texttt{GitHub repository}}. Unless stated otherwise, the nominal coverage level is fixed at $1-\alpha = 0.90$ and the tolerance $\tau$ of proposition~\ref{prop:prop1} is set at $\tau = 10^{-3}$.

\subsection{Academic benchmarks}
\label{sec4acad}

\subsubsection{2D parametric Poisson equation with moving source}
\label{sec41}

\noindent The field $u(\boldsymbol{\mu})$ defined on
$\Omega = (0,1)^2$ solves the parametric Poisson problem
\begin{equation}
\begin{cases}
-\Delta u(\boldsymbol{\mu}) = f(\boldsymbol{\mu}) & \text{in } \Omega, \\
u(\boldsymbol{\mu}) = 0 & \text{on } \partial\Omega,
\end{cases}
\end{equation}
where $\boldsymbol{\mu} = (\mu_0,\mu_1)\in[0,1]^2$ with $\pi_{\bm{\mu}}=\mathcal{U}[0,1]^{\otimes 2}$, and the source term is a moving Gaussian bump,
\begin{equation}
f(x,y;\boldsymbol{\mu})
= \exp\!\left(
    -\frac{(x-\mu_0)^2+(y-\mu_1)^2}{0.05}
  \right).
\end{equation}
The spatial discretization uses a five-point finite-difference Laplacian \citep{Quarteroni2008} on an interior grid of size $N = 100 \times 100 = 10^4$. The design of experiments consists of $n = 10^3$ parameter samples drawn uniformly in $[0,1]^2$, split into $n_{\mathrm{train}} = 600$, $n_{\mathrm{cal}} = 200$, and $n_{\mathrm{test}} = 200$. The KLE expansion retains $r = 13$ modes.

\begin{figure}[ht!]
\centering
\includegraphics[width=\linewidth]{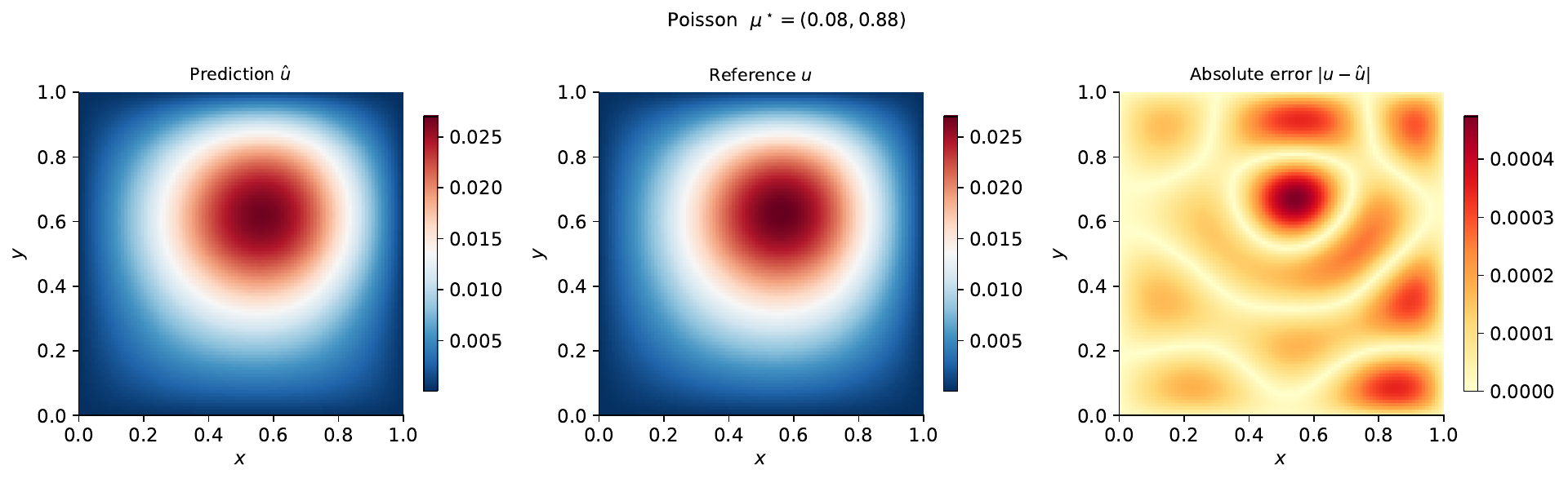}
\caption{%
    \textbf{Poisson surrogate prediction quality.}
    From left to right: NIROM prediction $\widehat u(\bm\mu^\star)$, high-fidelity reference $u(\bm\mu^\star)$, and absolute pointwise error for the representative test parameter $\bm\mu^\star = (0.08, 0.88)$. 
}
\label{fig:poisson_pred}
\end{figure}

\noindent Figure~\ref{fig:poisson_pred} shows the surrogate
prediction, the high-fidelity reference, and the pointwise absolute error for a representative test parameter $\bm\mu^\star = (0.08,\,0.88)$. With $r = 13$ modes, the NIROM reproduces the smooth response field essentially exactly, with a mean test $R^2$ of $0.993$ and a peak pointwise error below $5\times 10^{-4}$, i.e.\ under $2\%$ of the field maximum. The residual error exhibits the characteristic oscillatory imprint of the truncated KLE modes rather than any localized structural defect.

\begin{figure}[ht!]
\centering
\includegraphics[width=\linewidth]{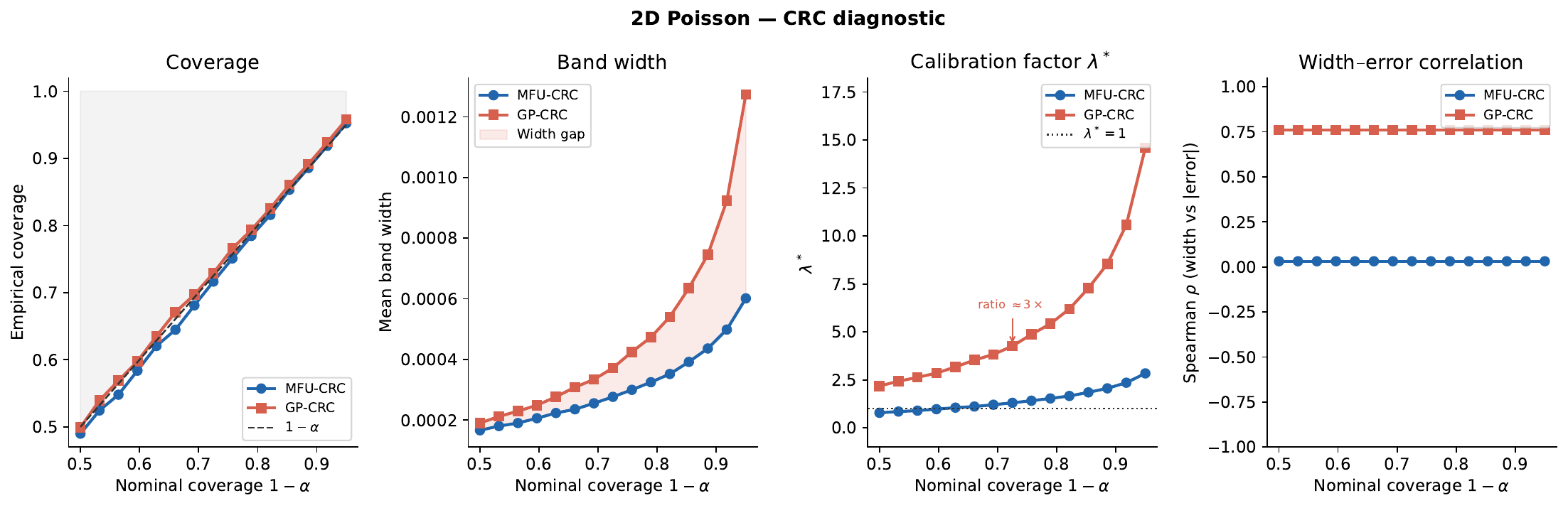}
\caption{%
    \textbf{Poisson - CRC diagnostics.}
    From left to right: empirical coverage vs nominal level $1-\alpha$; mean pointwise band width; calibration factor $\lambda^*$;  width-error Spearman correlation. MFU-CRC in blue, GP-CRC in orange.%
}
\label{fig:poisson_crc}
\end{figure}

\noindent The CRC diagnostics are collected in Figure~\ref{fig:poisson_crc}. Both normalizers track the diagonal in the coverage panel, as guaranteed by the distribution-free theory of step \stepCRC, but they differ in efficiency and adaptivity. The MFU bands are uniformly narrower, roughly half the width of the GP bands at $1-\alpha = 0.90$. The MFU calibration factor stays close to unity ($\lambda^*_{\mathrm{MFU}} = 2.06$ at $0.90$), so the perturbative variance is already an approximately valid predictive variance up to a factor of two, whereas the GP normalizer climbs from $\approx 2$ to nearly $15$ across the range of nominal levels ($\lambda^*_{\mathrm{GP}} = 8.55$ at $0.90$, a $4.1\times$ ratio). The width--error correlation panel reveals a complementary picture: on this smooth benchmark the GP posterior variance, although larger in scale than needed, is well correlated with the spatial error pattern ($\rho_{\text{GP}} \approx 0.76$), while the MFU width is nearly uncorrelated ($\rho_{\text{MFU}} \approx 0.03$). In other words, on problems where the GP regression error dominates, the GP variance carries the correct \emph{shape} and the perturbative variance the correct \emph{scale}; we return to this complementarity in the perspectives.

\subsubsection{1D parametric linear advection equation}
\label{sec42}
\noindent We solve the 1D linear advection equation on the periodic domain $\Omega = (0,1)$ with:
\begin{equation}
u_t + c(\mu)\,u_x = 0,
\qquad x\in(0,1),\ t\in[0,T],
\label{eq:advection}
\end{equation}
where the scalar parameter $\mu\in[0,1]$ controls the wave speed,
\begin{equation}
c(\mu) = c_0 + c_{\mathrm{scale}}\,\mu,
\end{equation}
and the initial condition is a periodic Gaussian bump,
\begin{equation}
u(0,x;\mu) = A\exp\!\left( -\frac{d(x,x_c)^2}{2\ell_{c}^{2}} \right), \qquad d(x,x_c) = \bigl((x - x_c + \tfrac{1}{2})\bmod 1\bigr) - \tfrac{1}{2},
\end{equation}
with $c_0 = 1$, $c_{\mathrm{scale}} = 1$, $x_c = 0.3$, $\ell_{c} = 0.03$, $A = 1$, and $T = 1.0$. The exact solution is available in closed form, $u(t,x;\mu) = u_0((x - c(\mu)\,t)\bmod 1)$, so no numerical discretization error is introduced. The spatial domain is discretized with $N_x = 200$ cells and $N_t = 200$ equispaced snapshot times, giving $N = N_x N_t = 4\times 10^4$. We use $n = 400$ parameter samples with $\mu \sim \mathcal{U}[0,1]$, split into $n_{\mathrm{train}} = 240$, $n_{\mathrm{cal}} = 80$, $n_{\mathrm{test}} = 80$. The KLE expansion retains $r = 3$ modes.

\noindent This benchmark is specifically designed to stress the POD-based ROM: the solution manifold consists of pure translates of the initial bump at different speeds, whose Kolmogorov $r$-width decays only algebraically \citep{Ohlberger2016, Greif2019}. A small number of POD modes cannot accurately represent the full range of bump positions, so the basis-truncation error is large by construction and the model-form uncertainty is the dominant contribution to the prediction error. This makes the advection problem the most informative benchmark for the perturbative MFU framework, and also, as we shall see, the one that most clearly exposes its limitations.

\begin{figure}[ht!]
\centering
\includegraphics[width=\linewidth]{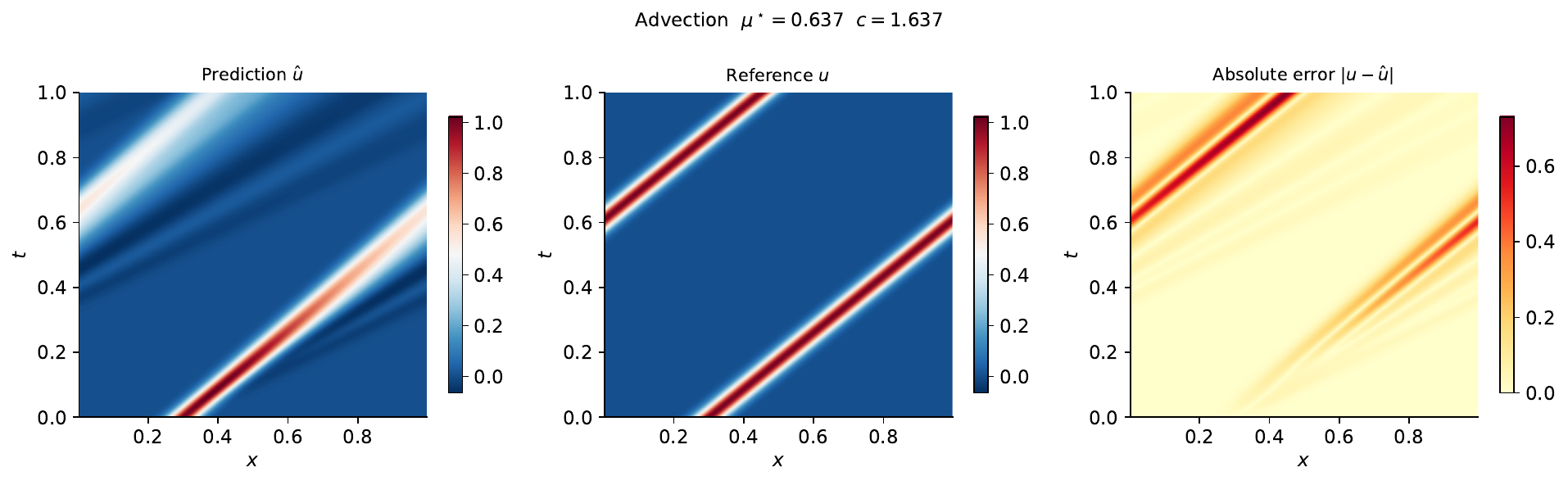}
\caption{%
    \textbf{Advection - surrogate prediction quality.}
    From left to right: NIROM prediction $\widehat u(\bm\mu^\star)$, high-fidelity reference $u(\bm\mu^\star)$, and absolute pointwise error $|u - \widehat u|$ shown in the $(t,x)$ space-time plane for the representative test parameter $\mu^\star = 0.637$ ($c = 1.637$). The 3-mode POD basis cannot represent the continuum of bump translates: the prediction smears the front into oscillatory lobes and the error, concentrated along the characteristic line, reaches a substantial fraction of the bump amplitude. The basis-truncation (model-form) error dominates by construction.%
}
\label{fig:advection_pred}
\end{figure}

\noindent Figure~\ref{fig:advection_pred} confirms this design: for the representative parameter $\mu^\star = 0.637$ ($c = 1.637$) the 3-mode reconstruction smears the sharp travelling bump into a sequence of oscillatory lobes, and the absolute error, concentrated along the characteristic line $x = x_c + c(\mu^\star)t \bmod 1$, reaches up to $70\%$ of the unit bump amplitude. The mean test $R^2$ of $0.353$ shows that the deterministic NIROM is, by construction, a poor pointwise predictor, and the value of the surrogate lies entirely in the quality of its uncertainty bands.

\begin{figure}[ht!]
\centering
\includegraphics[width=\linewidth]{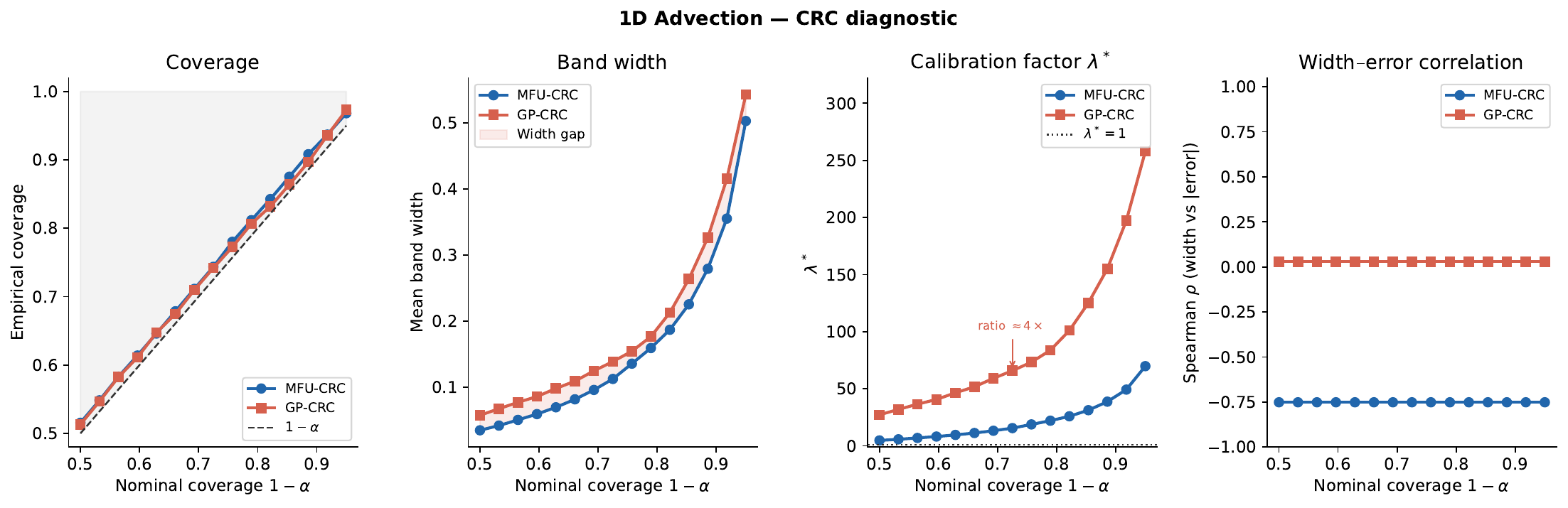}
\caption{%
    \textbf{Advection - CRC diagnostics.}
    Panels as in Figure~\ref{fig:poisson_crc}. Coverage holds despite severe misspecification; the $\lambda^*$ reduction from GP to MFU is $4.0\times$, the largest \emph{absolute} factors among the benchmarks. The width-error correlation is negative for MFU ($\rho\approx -0.75$): the perturbative variance captures the error scale but not its concentration along the characteristics, a structural limit of first-order perturbations of a linear basis.%
}
\label{fig:advection_crc}
\end{figure}

\noindent The CRC diagnostics in Figure~\ref{fig:advection_crc} show that both normalizers retain exact marginal coverage even in this severely misspecified regime. This is again the central robustness property of the conformal wrapper. The absolute scale mismatch of the GP posterior variance is the largest of all benchmarks: $\lambda^*_{\mathrm{GP}}$ grows from $\approx 27$ at $1-\alpha = 0.5$ to above $250$ at $0.95$ ($\lambda^*_{\mathrm{GP}} = 155$ at $0.90$), while the MFU calibration factor, though also large in absolute terms ($\lambda^*_{\mathrm{MFU}} = 38.8$ at $0.90$), remains a factor $4.0\times$ smaller. That the MFU factor is itself far from unity here reflects the severity of the misspecification: even the perturbed basis, being a small rotation of the retained subspace, only partially reaches the truncated complement where the advection error lives. Nonetheless the same mechanism as in the other benchmarks is at work --- the GP posterior variance, built from the regression residuals of the retained modal coefficients, is structurally blind to that complement, while the Stiefel perturbation of step \stepPERT\ probes it, yielding the consistent $4\times$ reduction in calibration factor.\\

\noindent The width-error correlation panel, however, reports a negative result and it delimits the scope of the method. Neither normalizer achieves a positive pointwise correlation in this transport-dominated regime, and the MFU width is anticorrelated with the error ($\rho_{\text{MFU}}\approx -0.75$, versus $\rho_{\text{GP}}\approx 0$). The perturbative variance correctly reduces the overall magnitude of the required correction but misplaces the uncertainty spatially: the basis perturbation spreads it across the support of the retained modes, while the true error concentrates on the thin characteristic band that those modes fail to represent at all. This behaviour is not an artifact of the present implementation but a structural limitation of any first-order perturbation of a linear basis facing a slow $r$-width manifold. The perturbed basis remains a small rotation of a subspace that does not contain the travelling front, so no amount of tilting within that neighbourhood can localise an error living outside it. Overcoming it requires leaving the linear setting altogether, and we retain this as an open problem. It is discussed further in Section~\ref{sec5}.

\subsubsection{1D viscous Burgers equation}
\label{sec43}

\noindent We solve the viscous 1D Burgers equation on $\Omega = (0,1)$ with periodic boundary conditions,
\begin{equation}
u_t + \left(\frac{u^2}{2}\right)_x = \nu\, u_{xx},
\qquad x\in(0,1),\ t\in[0,T],
\label{eq:burgers}
\end{equation}
with scalar parameter $\mu\in[0,1]$ controlling both the viscosity and the initial condition:
\begin{align}
\nu(\mu) &= \nu_{\min} + \mu\,(\nu_{\max}-\nu_{\min}), \\
u(0,x;\mu)
  &= \exp\!\left(-\frac{(x-x_c(\mu))^2}{2\ell_{c}^{2}}\right),
  \qquad x_c(\mu) = x_{c,0} + s\,(\mu - \tfrac{1}{2}),
\end{align}
so that $\mu$ simultaneously shifts the initial bump and changes the dissipation rate. We take $\nu_{\min} = 10^{-3}$, $\nu_{\max} = 10^{-2}$, $x_{c,0} = 1/2$, $s = 0.4$, $\ell_{c} = 0.05$, and $T = 1.0$, with $\mu\sim\mathcal{U}[0,1]$. The spatial domain is discretized with a uniform finite-volume mesh of $N_x = 200$ cells; the convective flux uses a Rusanov (local Lax-Friedrichs) scheme and time integration uses SSP-RK2 \citep{Gottlieb2001} with $\mathrm{CFL} = 0.4$. Snapshots are stored at $N_t = 200$ equispaced times, giving $N = N_x N_t = 4\times 10^4$. We use $n = 400$ parameter samples, split into $n_{\mathrm{train}} = 240$, $n_{\mathrm{cal}} = 80$, $n_{\mathrm{test}} = 80$. The KLE expansion retains $r = 3$ modes.

\begin{figure}[ht!]
\centering
\includegraphics[width=\linewidth]{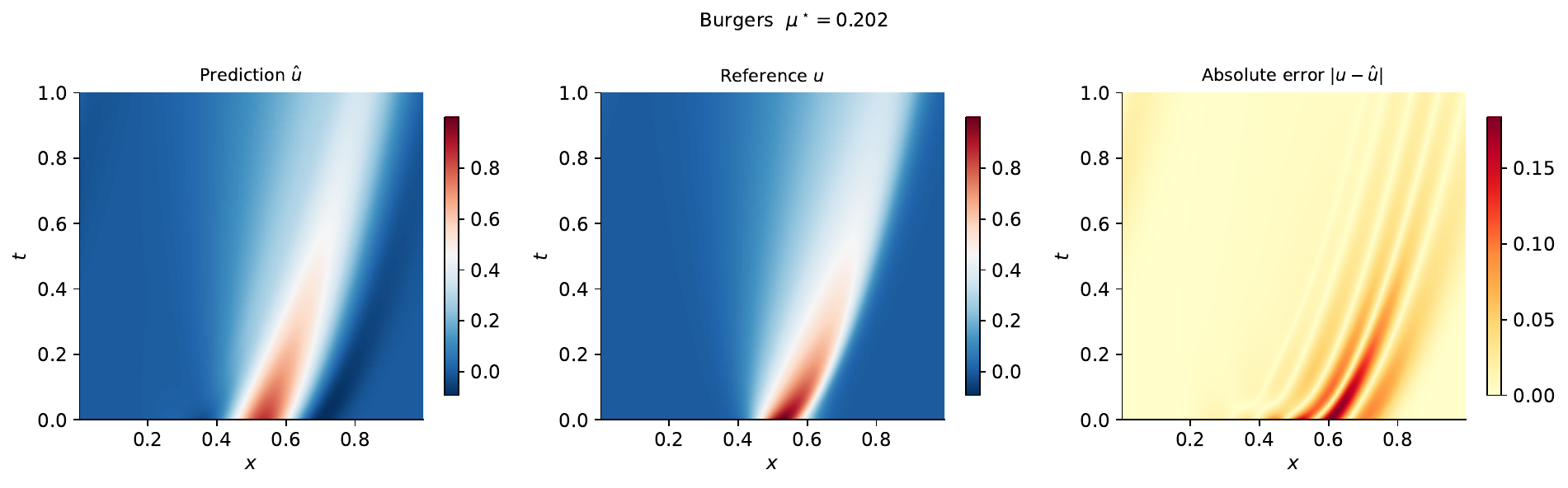}
\caption{%
    \textbf{Burgers - surrogate prediction quality.}
    From left to right: NIROM prediction $\widehat u(\bm\mu^\star)$, high-fidelity reference $u(\bm\mu^\star)$, and absolute pointwise error $|u - \widehat u|$ in the $(t,x)$ space-time plane for the representative test parameter $\mu^\star = 0.202$. The surrogate performs well almost everywhere; residual errors  concentrate in thin streaks along the steepening front at early times, where the 3-mode POD approximation is least accurate. This localised model-form error is precisely the signal that the perturbative MFU variance captures and the GP posterior variance
    misses.%
}
\label{fig:burgers_pred}
\end{figure}

\noindent The Burgers benchmark interpolates between the two previous regimes: the viscous dissipation restores partial compressibility of the solution manifold (mean test $R^2 = 0.849$ with only three modes), while the steepening front retains a localized structure that the linear basis struggles with. Figure~\ref{fig:burgers_pred} illustrates this for the representative parameter $\mu^\star = 0.202$ (a low-viscosity draw) $\nu \approx 2.8\times 10^{-3}$, hence a comparatively sharp front. The surrogate reproduces the global decay of the bump, and the absolute error, peaking around $15\%$ of the field maximum, is organized in thin streaks tracking the wave front at early times before viscosity smooths it out.

\begin{figure}[ht!]
\centering
\includegraphics[width=\linewidth]{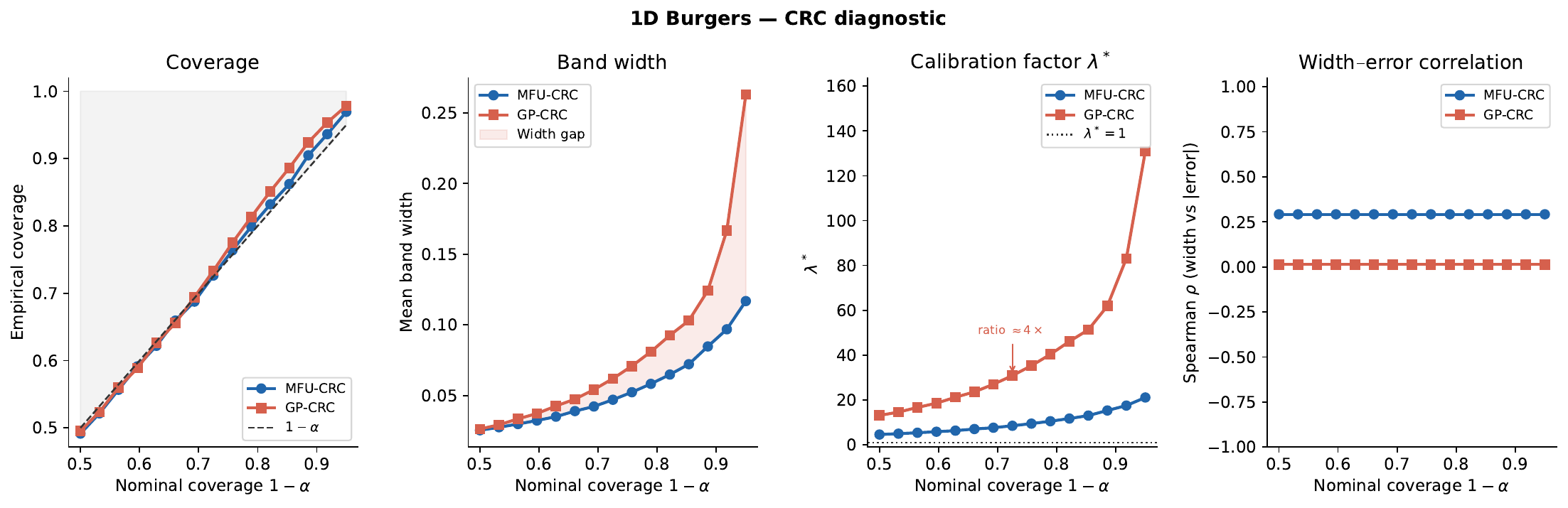}
\caption{%
    \textbf{Burgers - CRC diagnostics.} Panels as in Figure~\ref{fig:poisson_crc}. Both normalizers cover exactly; the $\lambda^*$ reduction from GP to MFU is $4.0\times$ with the band-width gap widening sharply above $1-\alpha = 0.85$. The MFU width is positively correlated with the local error ($\rho\approx 0.3$) whereas the GP width is uninformative,  showing that the perturbative variance captures both the scale and, partially, the location of the front-induced error.%
}
\label{fig:burgers_crc}
\end{figure}

\noindent The CRC diagnostics (Figure~\ref{fig:burgers_crc}) display the most favourable regime for the method. Coverage is exact for both normalizers, but the GP bands inflate sharply at high nominal levels. Indeed, at $1-\alpha = 0.95$ they are more than twice as wide as the MFU bands. The conformal correction must multiply an underdispersed variance by $\lambda^*_{\mathrm{GP}}\approx 13$-$130$ to reach the input coverage. The MFU normalizer requires $\lambda^*_{\mathrm{MFU}} = 15.3$ at $0.90$ against $\lambda^*_{\mathrm{GP}} = 61.9$, a $4.0\times$ reduction. Crucially, the MFU width is here positively correlated with the absolute error ($\rho_{\mathrm{MFU}}\approx 0.30$ against $\rho_{\mathrm{GP}}\approx 0$). When the model-form error is localized but still partially representable in the perturbed basis, the Stiefel perturbation captures both its scale and, to a useful extent, its location. This is the regime in which the method delivers its full benefit.

\subsection{Industrial use-case: 2D rubber calendering process}
\label{sec44}

\noindent Calendering is a common manufacturing technique, used in particular for tire production, that aims to produce thin rubber films by compressing material between several counter-rotating cylindrical rolls. We focus on the 2D cross-section of one pair of counter-rotating rolls.

\begin{figure}[ht!]
\centering
\resizebox{\textwidth}{!}{%
\includegraphics[width=\linewidth]{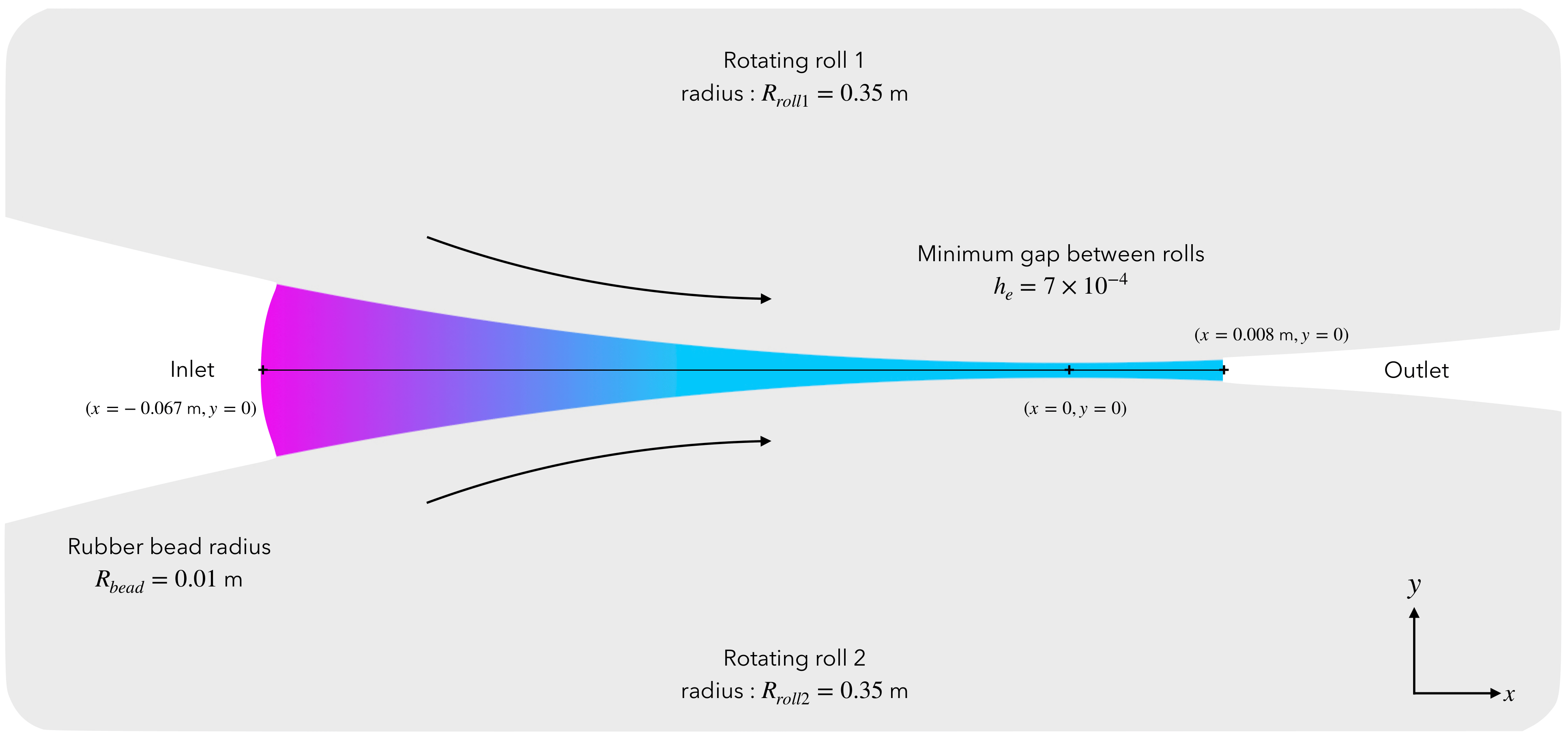}
}
\caption{%
    \textbf{2D calendering process - geometrical configuration.} The fluid enters a narrow gap between two counter-rotating rolls of radius $R_{\mathrm{roll}} = 0.35$\,m and is extruded through a minimum gap $h_e = 7\times 10^{-4}$\,m. The domain starts from a semicircular rubber bead of radius   $R_{\mathrm{bead}} = 0.01$\,m at the inlet and ends at the outlet plane. Geometrical parameters are held fixed; uncertainty is introduced  through two viscosity parameters $(K, n_{\mathrm{pl}})$ of the power-law constitutive model.%
}
\label{fig:calendering_schematic}
\end{figure}

\noindent The geometrical configuration is depicted in Figure~\ref{fig:calendering_schematic}. The fluid enters the narrow gap between two rolls (radius $R_{\mathrm{roll}} = 0.35$\,m) and is extruded through a minimum gap $h_e = 7\times 10^{-4}$\,m. The domain begins at a semicircular rubber bead of radius $R_{\mathrm{bead}} = 0.01$\,m at the inlet and ends at the outlet plane. Geometrical parameters are held fixed; uncertainty is introduced through two viscosity parameters. 

\noindent Rubber is modeled as an incompressible non-Newtonian fluid \citep{Nguyen2022} with a power-law viscosity:
\begin{equation}
\tau(\dot\gamma, K, n_{\mathrm{pl}}) = K\,\dot\gamma^{\,n_{\mathrm{pl}}-1},
\end{equation}
where $\dot\gamma$ is the shear rate, $K$ the consistency, and $n_{\mathrm{pl}}\in(0,1]$ the shear-thinning exponent. We write the exponent $n_{\mathrm{pl}}$ rather than $n$ to avoid any confusion with the size $n$ of the design of experiments introduced in Section~\ref{sec2}. The parameter vector $\bm\mu = (K,n_{\mathrm{pl}})\in\mathcal{X}\subset\mathbb{R}^2$ is sampled uniformly within physically admissible ranges for rubber manufacturing: $K \in [150,300]\times 10^3$ and $n_{\mathrm{pl}}\in[0.1,0.2]$. The high-fidelity solution is obtained from the in-house finite-element solver MEF++, co-developed with Université Laval \citep{Plasman-Thesis}. The output field of interest is the pressure field along the central streamline. The governing equations are the mass and momentum conservation laws:
\begin{equation}
\begin{cases}
\nabla\cdot\bigl(\tau(\dot\gamma;\bm\mu)\,\mathrm{sym}(\nabla u)\bigr)
  = \nabla p, \\
\nabla\cdot(\rho u) = 0,
\end{cases}
\label{eq:mass_conservation}
\end{equation}
solved with a Newton method and a custom line search.

\begin{figure}[ht!]
\centering
\includegraphics[width=\linewidth]{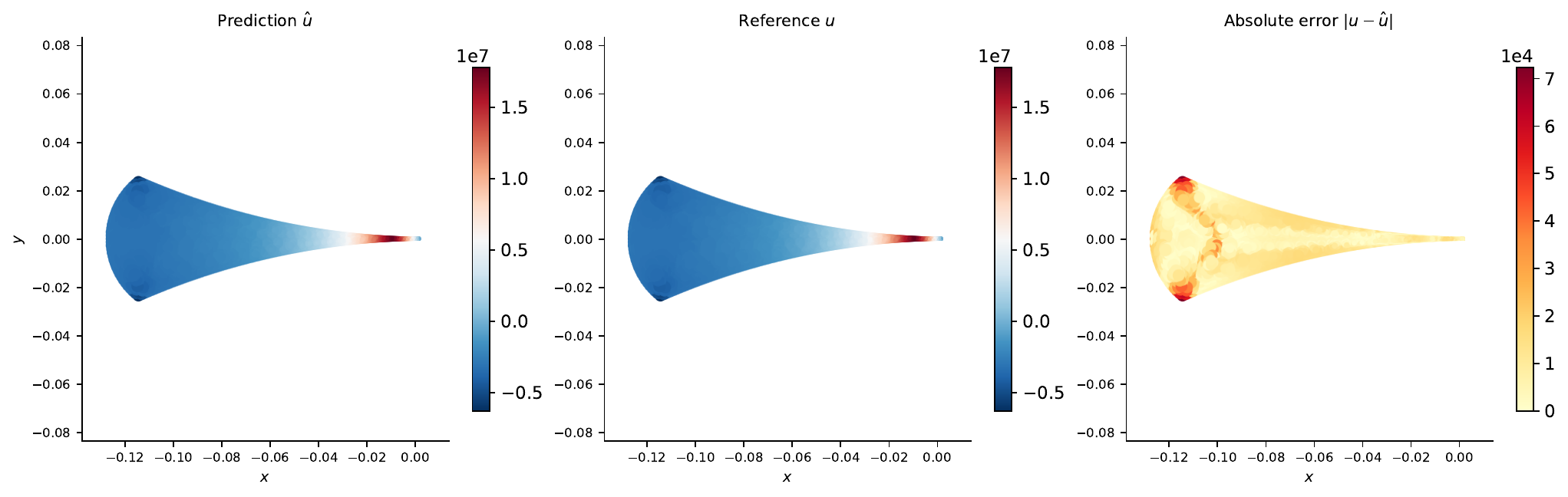}
\caption{%
    \textbf{Calendering - surrogate prediction quality.} From left to right: NIROM prediction $\widehat u(\bm\mu^\star)$, high-fidelity reference $u(\bm\mu^\star)$, and absolute pointwise error $|u - \widehat u|$ for a representative test parameter $\bm\mu^\star = (K^\star, n_{\mathrm{pl}}^\star)$. The surrogate accurately reproduces the pressure build-up toward the nip, with errors below half a percent of the field range; the residual error concentrates at the upstream corners of the inlet bead, where the constitutive nonlinearity and free-surface curvature are most pronounced.%
}
\label{fig:calendering_pred}
\end{figure}

\noindent Figure~\ref{fig:calendering_pred} shows the predicted and reference pressure fields on the unstructured finite-element domain. The NIROM captures the global pressure build-up toward the nip ($x\to 0$), where the field spans roughly $[-0.6,\,1.8]\times 10^{7}$\,Pa, with a pointwise error bounded by $7\times 10^{4}$\,Pa. The error is not uniformly distributed: it concentrates at the upstream corners of the rubber bead, where the free-surface curvature and the shear-thinning nonlinearity interact most strongly and where the retained modes are least expressive.

\begin{figure}[ht!]
\centering
\includegraphics[width=\linewidth]{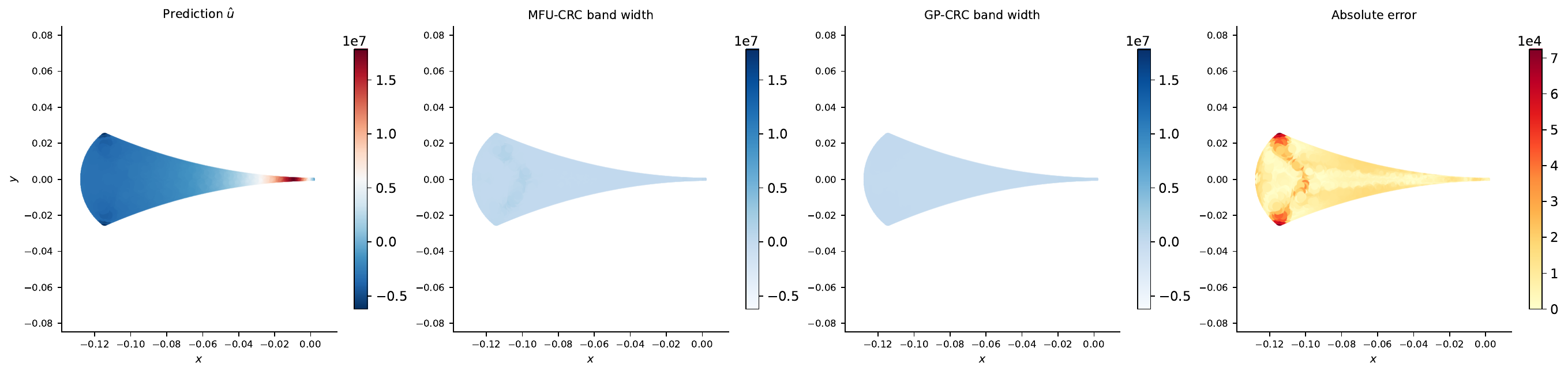}
\caption{%
    \textbf{Calendering - spatial structure of the calibrated bands.} From left to right: NIROM prediction, MFU-CRC pointwise band width, GP-CRC pointwise band width, and absolute error, for a representative test parameter at $1-\alpha = 0.90$. The MFU band exhibits a mild spatial modulation around the inlet bead inherited from the basis perturbation, while the GP band is nearly homogeneous; the corner-localised error peaks are not seen by either normalizer.%
}
\label{fig:calendering_bands}
\end{figure}

\noindent Figure~\ref{fig:calendering_bands} compares the spatial structure of the calibrated band widths. The MFU band inherits a mild spatial modulation from the perturbed basis, with slightly inflated widths near the bead region, whereas the GP band is essentially homogeneous over the domain; neither, however, fully resolves the corner-concentrated error pattern visible in the rightmost panel.

\begin{figure}[ht!]
\centering
\includegraphics[width=\linewidth]{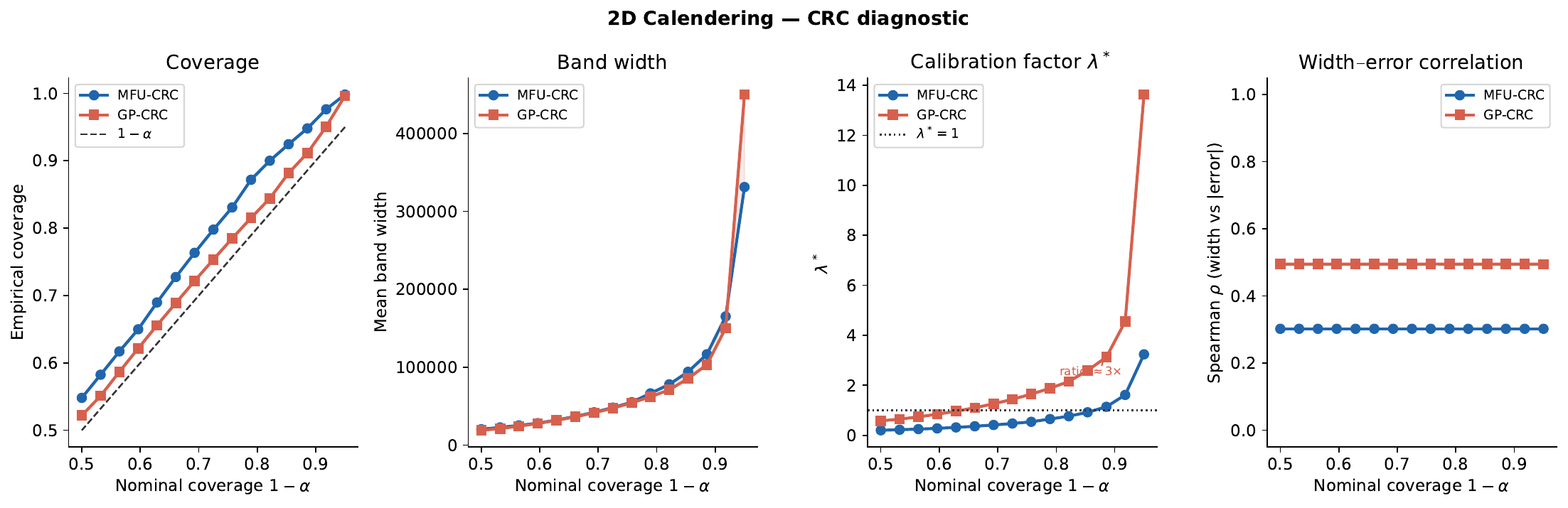}
\caption{%
    \textbf{Calendering - CRC diagnostics.}
    Panels as in Figure~\ref{fig:poisson_crc}. Both procedures cover, with MFU-CRC mildly conservative; the $\lambda^*$ reduction is $2.7\times$, the smallest of the benchmarks, and the GP width is the better localised ($\rho\approx 0.5$), consistent with regression error dominating on this compressible problem.%
}
\label{fig:calendering_crc}
\end{figure}

\noindent The CRC diagnostics (Figure~\ref{fig:calendering_crc}) show a behaviour distinct from the academic benchmarks, illustrating the value of running both normalizers in practice. This is the only benchmark where the surrogate is near-exact with very few modes ($R^2 = 0.999$ with $r = 2$): the pressure manifold is highly compressible, so the residual is dominated by regression noise rather than basis truncation. Both procedures satisfy the coverage guarantee, with MFU-CRC exhibiting mild conservative over-coverage. The MFU calibration factor sits close to and partly below unity ($\lambda^*_{\mathrm{MFU}} = 1.14$ at $0.90$), while the GP factor, better behaved here than on the PDE benchmarks, reaches $\lambda^*_{\mathrm{GP}} = 3.13$ at the same level.
Consistently with the regression-dominated regime, the GP width is the better correlated with the local error ($\rho_{\text{GP}}\approx 0.50$ versus $0.30$ for MFU). The overall picture matches the mechanism identified throughout this section: the perturbative MFU variance is the appropriate normalizer when basis truncation dominates the error budget, while the GP posterior variance remains competitive when the manifold is easy to reduce and regression error dominates, with the conformal
wrapper guaranteeing validity in either case.

\section{Conclusion and perspectives}
\label{sec5}

\noindent We have presented a probabilistic framework for quantifying model-form uncertainty in non-intrusive POD-GP reduced-order models. The proposed methodology combines random perturbations of the reduced basis on the Stiefel manifold with a first-order transport approximation of the Gaussian-process posterior, resulting in a closed-form decomposition of predictive uncertainty into two complementary contributions: the uncertainty associated with learning the reduced coordinates and the uncertainty induced by the reduced basis itself. This construction avoids retraining the Gaussian processes for each perturbed basis and therefore provides an inexpensive uncertainty propagation strategy that remains fully non-intrusive.\\

\noindent A second contribution of this work is the integration of this uncertainty model within a conformal risk control framework. Rather than relying solely on Gaussian credibility intervals, the proposed approach produces prediction bands with finite-sample, distribution-free guarantees on the expected coordinate-wise miscoverage. Beyond providing calibrated prediction sets, the conformal calibration factor offers an interpretable quantitative diagnostic of the underlying uncertainty estimate. It therefore constitutes a convenient tool for comparing different uncertainty models independently of the calibration procedure itself.\\

\noindent The numerical experiments illustrate two complementary aspects of the methodology. First, incorporating basis perturbations consistently reduces the amount of conformal correction required to obtain reliable prediction bands compared with using the Gaussian-process posterior variance alone. This behaviour is particularly pronounced on problems where basis truncation contributes significantly to the prediction error. Second, the experiments also highlight the intrinsic limitations of the perturbative approximation. For transport-dominated problems such as the linear advection equation, the perturbative variance captures the global magnitude of the model-form uncertainty but does not accurately localize the prediction error along propagating characteristics. This behaviour reflects the limited expressive power of linear reduced spaces rather than the conformal calibration itself. Conversely, on problems exhibiting smoother solution manifolds, such as the Poisson and calendering benchmarks, or moderately nonlinear dynamics such as the viscous Burgers equation, the perturbative uncertainty remains spatially informative and leads to efficient calibrated prediction bands.\\

\noindent More generally, the proposed framework should be viewed as a principled methodology for enriching Bayesian surrogate uncertainty with an explicit structural contribution arising from the reduced basis. The law-of-total-covariance decomposition~\eqref{eq:total_variance} naturally separates regression-induced and basis-induced uncertainties while remaining computationally tractable through the transport approximation. The conformal calibration layer is complementary to this construction: it provides rigorous finite-sample guarantees while simultaneously serving as an objective validation tool for assessing the quality of different uncertainty models.\\

\noindent Several directions naturally emerge from this work. On the theoretical side, developing higher-order transport approximations could extend the validity of the framework beyond the small-perturbation regime considered here. On the modeling side, extending the perturbation framework to nonlinear latent representations, such as autoencoder-based reduced-order models, appears particularly promising for problems whose solution manifolds cannot be accurately represented by linear subspaces. Finally, the present methodology naturally accommodates alternative structural uncertainty models, making it an attractive foundation for future developments combining probabilistic reduced-order modeling, Bayesian inference, and distribution-free uncertainty quantification.

\section*{Acknowledgements}
\noindent This research was conducted as part of a postdoctoral research grant obtained by the first author from the chair of Industrial Data-Analytics \& Machine Learning (IDAML) of Centre Borelli (ENS Paris-Saclay). The first author would like to thank Dr. Vincent Blot for useful discussions about the present work.

\appendix
\section{Proof of Proposition~\ref{prop:prop1}}
\label{app:proof1}

\noindent
Let
\begin{equation}
M_{\varepsilon} = \bm{\Phi}_{*} + \varepsilon\bm{\Phi}_{\perp}W_{\xi},
\end{equation}
where $W_{\xi}\in\mathbb{R}^{r_{\perp}\times r}$ is a Gaussian random matrix. Let
\begin{equation}
M_{\varepsilon} = Q_{\varepsilon}^{\rm raw} R_{\varepsilon}^{\rm raw}
\end{equation}
denote its reduced $QR$ factorization. To remove the sign ambiguity of the decomposition, introduce
\begin{equation}
D_{\varepsilon} = \text{diag}\!\left(\text{sign}\bigl(\text{diag}(R_{\varepsilon}^{\rm raw})\bigr)\right),
\end{equation}
where zero diagonal entries are replaced by $1$. The perturbed basis defined as in Eq.~\eqref{eq:perturbation_basis} is:
\begin{equation}
\bm{\Phi}(\xi) = Q_{\varepsilon}^{\rm raw}D_{\varepsilon}.
\end{equation}
Define the associated upper-triangular factor:
\begin{equation}
C_{\varepsilon} = D_{\varepsilon} R_{\varepsilon}^{\rm raw}.
\end{equation}
Since $D_{\varepsilon}^{2}=I_r$, one has
\begin{equation}
M_{\varepsilon}
=
\bm{\Phi}(\xi)\,C_{\varepsilon},
\end{equation}
where $C_{\varepsilon}$ is upper triangular with strictly positive diagonal
entries.

\medskip

\noindent
Using
\begin{equation}
\bm{\Phi}_{*}^{\top}\bm{\Phi}_{*}=I_r,
\qquad
\bm{\Phi}_{*}^{\top}\bm{\Phi}_{\perp}=0,
\qquad
\bm{\Phi}_{\perp}^{\top}\bm{\Phi}_{\perp}=I_{r_{\perp}},
\end{equation}
we obtain
\begin{align}
M_{\varepsilon}^{\top}M_{\varepsilon}
&=
\bm{\Phi}_{*}^{\top}\bm{\Phi}_{*}
+
\varepsilon
\left(
\bm{\Phi}_{*}^{\top}\Delta_{\xi}
+
\Delta_{\xi}^{\top}\bm{\Phi}_{*}
\right)
+
\varepsilon^{2}\Delta_{\xi}^{\top}\Delta_{\xi}
\\
&=
I_r+\varepsilon^{2}W_{\xi}^{\top}W_{\xi},
\label{eq:qr_gram_matrix}
\end{align}
since
\begin{equation}
\Delta_{\xi} = \bm{\Phi}_{\perp}W_{\xi},
\qquad \bm{\Phi}_{*}^{\top}\Delta_{\xi}=0,
\end{equation}
and
\begin{equation}
\Delta_{\xi}^{\top}\Delta_{\xi} = W_{\xi}^{\top} \bm{\Phi}_{\perp}^{\top}\bm{\Phi}_{\perp} W_{\xi} = W_{\xi}^{\top}W_{\xi}.
\end{equation}

\noindent On the other hand,
\begin{equation}
M_{\varepsilon} = \bm{\Phi}(\xi)\,C_{\varepsilon},
\end{equation}
implies
\begin{equation}
M_{\varepsilon}^{\top}M_{\varepsilon} = C_{\varepsilon}^{\top}C_{\varepsilon},
\end{equation}
so that $C_{\varepsilon}$ is the unique Cholesky factor of
\begin{equation}
I_r+\varepsilon^{2}G_{\xi},
\qquad
G_{\xi}=W_{\xi}^{\top}W_{\xi}.
\end{equation}
Since the Cholesky factor depends analytically on its argument in a neighbourhood of the identity, it admits the expansion:
\begin{equation}
C_{\varepsilon} = I_r+\varepsilon^{2}T_{\xi}+O(\varepsilon^{4}),
\end{equation}
where $T_{\xi}$ is upper triangular. Substituting into
\begin{equation}
C_{\varepsilon}^{\top}C_{\varepsilon} = I_r+\varepsilon^{2}G_{\xi}
\end{equation}
gives
\begin{equation}
T_{\xi}+T_{\xi}^{\top} = G_{\xi},
\end{equation}
and therefore:
\begin{equation}
    T_{\xi} = \frac{1}{2}W_{\xi}^{\top}W_{\xi}.
\end{equation}
The overlap matrix is
\begin{equation}
R_{\varepsilon}(\xi) =
\bm{\Phi}_{*}^{\top}\bm{\Phi}(\xi).
\end{equation}
Since
\begin{equation}
\bm{\Phi}(\xi) = M_{\varepsilon}C_{\varepsilon}^{-1},
\end{equation}
one has:
\begin{equation}
R_{\varepsilon}(\xi) = \bm{\Phi}_{*}^{\top} M_{\varepsilon} C_{\varepsilon}^{-1} = \left( I_r+\varepsilon\bm{\Phi}_{*}^{\top}\Delta_{\xi} \right) C_{\varepsilon}^{-1} = C_{\varepsilon}^{-1},
\end{equation}
because $\bm{\Phi}_{*}^{\top}\Delta_{\xi}=0$. Finally, we get
\begin{equation}
C_{\varepsilon}^{-1} = I_r-\varepsilon^{2}T_{\xi}+O(\varepsilon^{4}),
\end{equation}
so that
\begin{equation}
R_{\varepsilon}(\xi) = I_r-\varepsilon^{2}T_{\xi}+O(\varepsilon^{4}).
\end{equation}
Taking Frobenius norms gives
\begin{equation}
\|R_{\varepsilon}(\xi)-I_r\|_F = \varepsilon^{2}\|T_{\xi}\|_F + O(\varepsilon^{4}),
\end{equation}
and therefore
\begin{equation}
\mathbb{E}_{\xi} \!\left[ \|R_{\varepsilon}(\xi)-I_r\|_F \right] = \varepsilon^{2} \mathbb{E}_{\xi} \!\left[ \|T_{\xi}\|_F \right] + O(\varepsilon^{4}).
\end{equation}
Hence, for a prescribed overlap tolerance $\tau>0$, choosing
\begin{equation}
\varepsilon^{*} = \left( \frac{2\tau} {\mathbb{E}_{\xi}\!\left[\|W_{\xi}^{\top}W_{\xi}\|_F\right]} \right)^{1/2}
\end{equation}
yields
\begin{equation}
\mathbb{E}_{\xi} \!\left[ \|R_{\varepsilon^{*}}(\xi)-I_r\|_F \right] = \tau+O\!\left((\varepsilon^{*})^{4}\right),
\end{equation}
which completes the proof.
\hfill$\square$

\section{Gaussian process regression}
\label{app:gp}
\noindent Let $\eta:\cX\to\mathbb{R}$ denote a scalar reduced coordinate obtained from the POD modes as in Eq.~\eqref{eq:kl_modes}. In the Gaussian process framework, $\eta$ is modeled as:
\begin{equation}
\eta(\bm{\mu}) = m(\bm{\mu}) + W(\bm{\mu}),
\end{equation}
where $m(\bm{\mu})$ is a deterministic mean function and $W$ is a centered Gaussian process with covariance kernel:
\begin{equation}
\mathrm{Cov}\big(W(\bm{\mu}),W(\bm{\mu}')\big)
= k_{\theta}(\bm{\mu},\bm{\mu}'),
\end{equation}
parameterized by hyperparameters $\theta$. Throughout this work, we consider a constant mean function $m(\bm{\mu})=\beta\in\R$. Given a design of experiments
\begin{equation}
\mathrm{DoE}_n = \left\{
\big(\bm{\mu}^{(i)},\eta(\bm{\mu}^{(i)})\big)
\right\}_{i=1}^{n},
\end{equation}
we introduce the observation vector:
\begin{equation}
\bm{\eta}_n = \big(\eta(\bm{\mu}^{(1)}),\ldots,\eta(\bm{\mu}^{(n)})\big)^\top,
\end{equation}
together with the covariance matrix $\bm{K}_n\in\mathbb{R}^{n\times n}$ defined by:
\begin{equation}
(\bm{K}_n)_{ij}=k_{\theta}(\bm{\mu}^{(i)},\bm{\mu}^{(j)}).
\end{equation}
For a new parameter value $\bm{\mu}\in\mathcal{X}$, we further define the covariance vector:
\begin{equation}
\bm{k}_n(\bm{\mu}) =\big(k_{\theta}(\bm{\mu},\bm{\mu}^{(1)}),\ldots,k_{\theta}(\bm{\mu},\bm{\mu}^{(n)})\big)^\top.
\end{equation}
Under the GP prior assumption, the joint distribution of $(\bm{\eta}_n,\eta(\bm{\mu}))$ is Gaussian. Conditioning with respect to the observations yields the classical Kriging predictor. The posterior mean is given by
\begin{equation}
\widehat{\eta}(\bm{\mu})=\beta+\bm{k}_n(\bm{\mu})^\top\bm{K}_n^{-1}(\bm{\eta}_n-\beta\bm{1}_n),
\end{equation}
while the associated posterior variance reads:
\begin{equation}
\widehat{s}^{\,2}(\bm{\mu})=k_{\theta}(\bm{\mu},\bm{\mu})-\bm{k}_n(\bm{\mu})^\top\bm{K}_n^{-1}\bm{k}_n(\bm{\mu}).
\end{equation}
We write this scalar posterior variance $\widehat{s}^{\,2}$ rather than $\widehat{\gamma}$, the latter being reserved throughout the paper for the $N\times N$ field covariance of Eq.~\eqref{eq:total_variance}. Collecting the $r$ scalar posterior variances on the diagonal yields the matrix $\widehat{\bm{\Sigma}}_{*}(\bm{\mu})$ used in Section~\ref{sec33}.
The hyperparameters $(\beta,\theta)$ are estimated by maximum likelihood, leading to the standard Kriging predictor \citep{Rasmussen2006}. In practice, a small nugget parameter $\varsigma^{2}>0$ is added to the diagonal of $\bm{K}_n$ to improve numerical stability and, when appropriate, account for a possible noise component in the observations.

\bibliographystyle{elsarticle-num}
\bibliography{mfu-nirom}
\end{document}